\documentclass[lettersize,journal]{IEEEtai}

\usepackage[colorlinks,urlcolor=blue,linkcolor=blue,citecolor=blue]{hyperref}
\usepackage{color,array}

\usepackage{graphicx}

\usepackage{graphicx}
\usepackage{pifont}
\usepackage{amsmath}
\usepackage{amssymb}
\usepackage{mathrsfs}

\usepackage{makecell}
\usepackage{float}
\usepackage[normalem]{ulem}
\useunder{\uline}{\ul}{}
\usepackage{multirow}
\usepackage{booktabs}
\usepackage{subcaption}
\usepackage{xcolor}
\usepackage[justification=raggedright,singlelinecheck=false]{caption}
\begin{document}

% \title{PM-DMNet: Pattern-Matching Dynamic Memory Network for Traffic Prediction} 
\title{
CoMemNet: A Continual Memory Network with Drift-Aware Sampling for Traffic Prediction
}
\author{Mei Wu, \IEEEmembership{Graduate Student Member, IEEE}, Wenchao Weng \IEEEmembership{Student Member, IEEE}, Wenxin Su, Wenjie Tang, Wei Zhou, \IEEEmembership{Member, IEEE}
\thanks{*corresponding author: Wenchao Weng.}
\thanks{Mei Wu is with the College of Computer Science, Shanghai Jiao Tong University, Shanghai 200240, China (e-mail: wumei5@sjtu.edu.cn).}
\thanks{Wenchao Weng and Wenxin Su are with the College of Computer Science and Technology, Zhejiang University of Technology, Hangzhou 310023, China (e-mail: 111124120010@zjut.edu.cn; suwenxin@zjut.edu.cn).}
\thanks{Wenjie Tang  and Wei Zhou are with the School of Automation, Nanjing University of Science and Technology, Nanjing 210094, China (e-mail: 15209822362@njust.edu.cn; weizhou@njust.edu.cn).}
% \thanks{Feng Xia is with School of Computing Technologies, RMIT University, Melbourne, VIC 3000, Australia (e-mail: f.xia@ieee.org).
% }
% \thanks{This paragraph will include the Associate Editor who handled your paper.}
}

% \markboth{Journal of IEEE Transactions on Artificial Intelligence, Vol. 00, No. 0, Month 2020}
% {First A. Author \MakeLowercase{\textit{et al.}}: Bare Demo of IEEEtai.cls for IEEE Journals of IEEE Transactions on Artificial Intelligence}

\maketitle
\begin{abstract}
Traffic sensor networks evolve as sensors are added and traffic distributions change, whereas most forecasting models assume a fixed node set and repeatedly retrain on all available data. We propose CoMemNet, a \emph{Continual Memory Network} for efficient prediction over evolving traffic sensor networks. CoMemNet uses an Online branch to adapt to the current period and an exponential-moving-average Target branch as a stable feature reference. A Wasserstein-based Drift Sampler compares node-wise Online--Target feature distributions and selects a limited set of drift-sensitive nodes for updating. A lightweight Node-Adaptive Temporal Memory Replay Buffer (TMRB-N) retains compact temporal states without repeatedly traversing all historical training data. The prediction backbone does not consume an adjacency matrix; sensor adjacency is used only to construct data and optionally expand the selected update set to a limited neighborhood. Experiments on three multi-period PeMS datasets include three-seed evaluation, strong static retraining and continual baselines, controlled sampling strategies, continual-learning metrics, robustness tests, and resource accounting. The results show that CoMemNet maintains stable prediction accuracy and efficient adaptation under bounded shared-node selection, achieving a better balance between historical knowledge preservation and current-period prediction performance. Meanwhile, as the evolving network expands, CoMemNet shows clearer accuracy and cumulative training-time advantages over current-period retraining baselines. The project page is available at
\href{https://meiwu5.github.io/CoMemNet/}
{\textcolor{blue}{https://meiwu5.github.io/CoMemNet/}}.

\end{abstract}

% \begin{IEEEImpStatement}
% The impact statement should not exceeed 150 words. This section offers an example that is expanded to have only and just 150 words to demonstrate the point. Here is an example on how to write an appropriate impact statement: Chatbots are a popular technology in online interaction. They reduce the load on human support teams and offer continuous 24-7 support to customers. However, recent usability research has demonstrated that 30\% of customers are unhappy with current chatbots due to their poor conversational capabilities and inability to emotionally engage customers. The natural language algorithms we introduce in this paper overcame these limitations. With a significant increase in user satisfaction to 92\% after adopting our algorithms, the technology is ready to support users in a wide variety of applications including government front shops, automatic tellers, and the gaming industry. It could offer an alternative way of interaction for some physically disable users.
% \end{IEEEImpStatement}


\begin{figure}[ht]

\end{figure}

\section{Introduction}

\IEEEPARstart{T}{he} task of traffic prediction is to forecast future traffic conditions based on historical traffic data obtained from sensors\cite{sensors1,sensors2,sensors3}. Its reliability and advancement have driven the development of traffic planning, intelligent transportation systems, and urban sustainability research. In recent years, combining non-topological traffic network modeling with different temporal learning methods has become a mainstream approach to effectively capture the spatio-temporal heterogeneous features of non-Euclidean graph structures, which has proven successful in addressing this issue\cite{GraphWaveNet,DCRNN,ASTGCN}.

Although spatio-temporal models have been widely applied in traffic flow prediction\cite{ASTGCN,DSANet}, most existing methods primarily focus on short-term forecasting (typically trained on approximately one month of data) and assume a static road network graph structure (i.e., neglecting topological changes due to urban expansion). However, in reality, traffic networks exhibit continuous expansion and evolution, with new road networks and nodes being added over time. As a result, both the underlying graph structure and flow patterns change over time\cite{expand,Trafficstream,TII1}. While retraining all previous training samples could address this issue, it would lead to substantial computational and storage overheads in large-scale data networks\cite{TVT_traffic_1,TVT_traffic_2,TVT_traffic_3}. Therefore, constructing a long-term streaming network that can adapt to the evolving and expanding traffic network has become a relatively new challenge in current research\cite{PECPM,STECK,TII2}.

\begin{figure}
    \centering
    \includegraphics[width=0.83\linewidth]{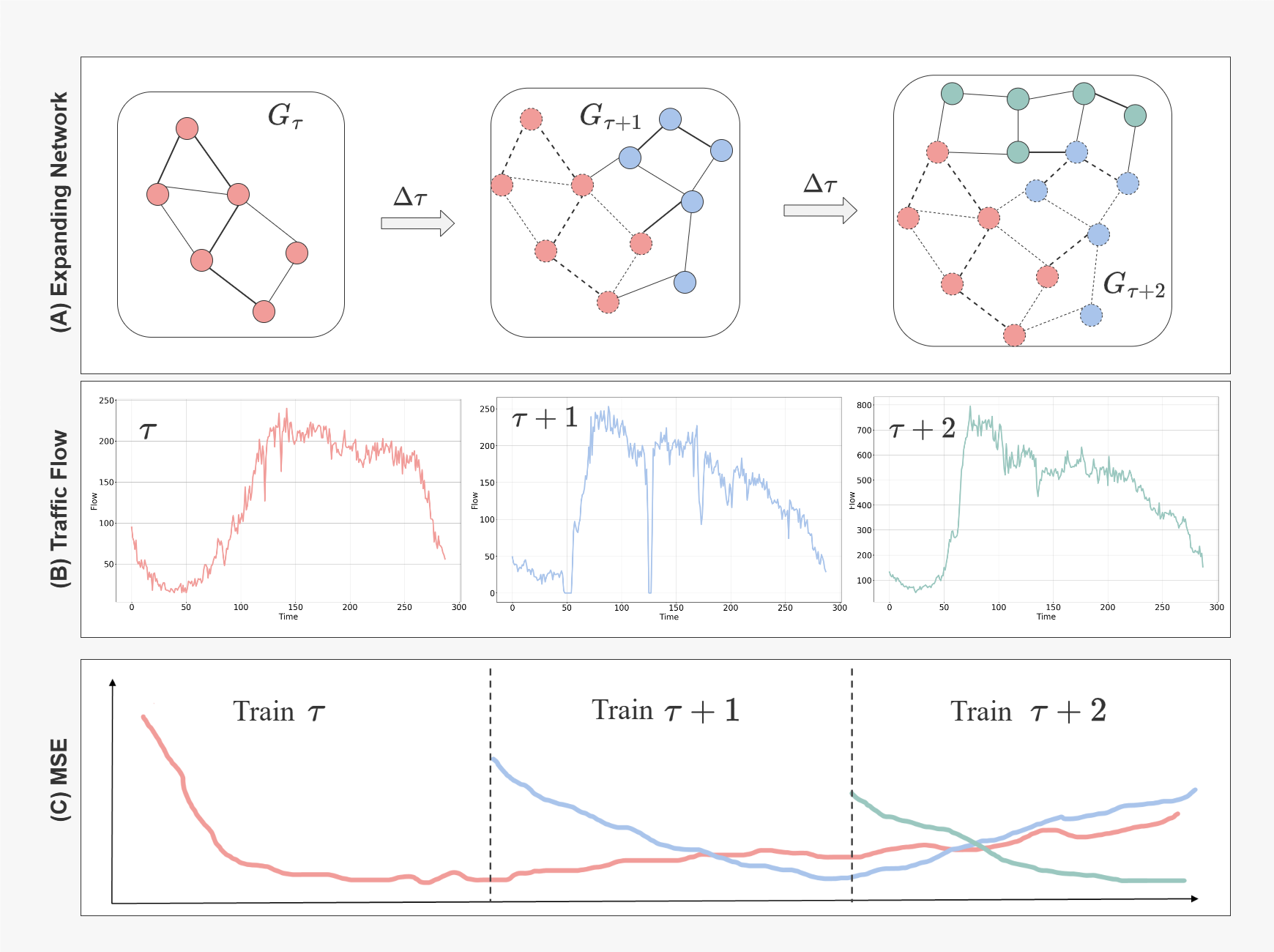}
    \caption{(A) Expansion of the traffic network structure (B) The traffic flow data for Station 40009 in District 4 of the PeMS system from August 1st, 2010 to 2012. (C) 
The prediction performance on old tasks deteriorates when training on new tasks.}
    \label{fig:motivation}
\end{figure}

% Continual learning\cite{CL1,CL2,CL3}, also known as lifelong learning, aims to address ongoing changes in traffic network topology and flow patterns through the expansion and consolidation of knowledge. Applying continuous learning to streaming traffic networks with spatio-temporal relationships requires overcoming the following three challenges:

To address the aforementioned challenge of dynamic network evolution, continual learning\cite{CL1,CL2,CL3} (also known as lifelong learning) provides an efficient incremental learning paradigm aimed at handling the continuous changes in traffic network topology and flow patterns through knowledge expansion and knowledge consolidation. Applying continual learning to spatio-temporal streaming traffic networks requires overcoming the following three challenges:

\textbf{Prediction over Evolving Sensor Sets.}
In the expanding network shown in Fig. \ref{fig:motivation}(A), sensors are added over time and the available node set changes across periods. Predictors tied to a fixed adjacency size require structural reconstruction when the sensor set grows. This motivates an adjacency-free prediction backbone whose parameters can be reused across periods. We emphasize that adjacency metadata may still support data construction and limited neighborhood expansion during updating; the predictor itself does not receive the adjacency matrix.

\textbf{Efficient Adaptation under Spatio-Temporal Drift.}
As illustrated in Fig. \ref{fig:motivation}(B) and (C), the traffic-flow distribution of the same node can change across periods. Updating all nodes and revisiting all historical data is costly, while updating too few nodes limits adaptation. We therefore use a slowly updated Target branch as a stable reference and compare its features with Online-branch features to identify nodes exhibiting large representation-distribution shifts.

\textbf{Growing Historical Storage and Computation.}
Repeated all-history retraining causes cumulative data access and training cost to grow with the number of periods. Storing complete historical neighborhoods also scales poorly as the sensor set expands. A compact temporal state and a budgeted update policy are therefore desirable for long-term continual prediction.

In this work, we propose \textbf{CoMemNet}, short for \emph{Continual Memory Network}. Its prediction backbone does not consume an explicit adjacency matrix, which allows the same predictor to operate on evolving sensor sets. Adjacency metadata is used only outside the predictor for data construction and limited $K_{\mathrm{hop}}$-hop expansion of the selected update set \textbf{(Challenge 1)}.

For efficient continual adaptation \textbf{(Challenge 2)}, the Online branch learns the current period while an exponential-moving-average Target branch provides a smoother feature reference. We introduce a Wasserstein-based Drift Sampler (DS) that ranks shared nodes by their Online--Target representation discrepancy and performs bounded shared-node selection. This is a practical adaptation-demand proxy, not a direct raw-data period-to-period distance and not a representation-level contrastive-learning objective.

To limit growing historical cost \textbf{(Challenge 3)}, the selected nodes interact with compact temporal states stored in a Node-Adaptive Temporal Memory Replay Buffer (TMRB-N). The buffer performs latent-state reuse through a gated update, avoiding repeated traversal of all historical raw sequences. Resource accounting is reported explicitly rather than describing full retraining as universally infeasible.

Our contributions can be summarized as follows.
\begin{itemize}
\item We propose CoMemNet, a Continual Memory Network with an adjacency-free prediction backbone for evolving traffic sensor sets. We explicitly separate adjacency-free prediction from topology-assisted update-set expansion.

\item We design an Online--Target feature-comparison framework. The Online branch adapts to the current period, whereas the EMA Target branch provides a smoother reference for estimating node-wise representation drift.

\item We introduce a Wasserstein-based Drift Sampler and a lightweight Node-Adaptive Temporal Memory Replay Buffer (TMRB-N). Together they support bounded shared-node selection and compact temporal-state reuse without repeatedly training on the complete historical record.

\item We contribute two curated multi-period PeMS datasets and perform extensive evaluation with three random seeds, strong retrained and continual baselines, sampler and component controls, task-wise continual-learning metrics, robustness tests, and measured time, memory, and historical-data access.
\end{itemize}

\section{Related Work}
\subsection{Traffic Flow Prediction}

Traffic forecasting has been studied with graph neural networks, attention models, and lightweight embedding-based predictors. STGNNs such as PM-DMNet\cite{PM-DMNet}, TSHDNet\cite{TSHDNet}, and D2STGNN\cite{D2STGNN} model spatial dependence with graph operations, while attention-based methods such as STD-MAE\cite{STD-MAE}, STWave\cite{STWave}, and PDFormer\cite{PDFormer} improve long-range temporal or spatial modeling. Recent large spatio-temporal language models\cite{PromptCast,ST-LLM,GATGPT} further broaden the forecasting paradigm but can be costly to update repeatedly. Lightweight models such as STID\cite{STID} show that strong static forecasting accuracy can also be obtained with simple embeddings. These methods mainly assume a fixed evaluation period or retraining protocol; our focus is efficient adaptation when the sensor set and traffic distribution evolve across periods.
\begin{figure*}
    \centering
    \includegraphics[width=0.95\linewidth]{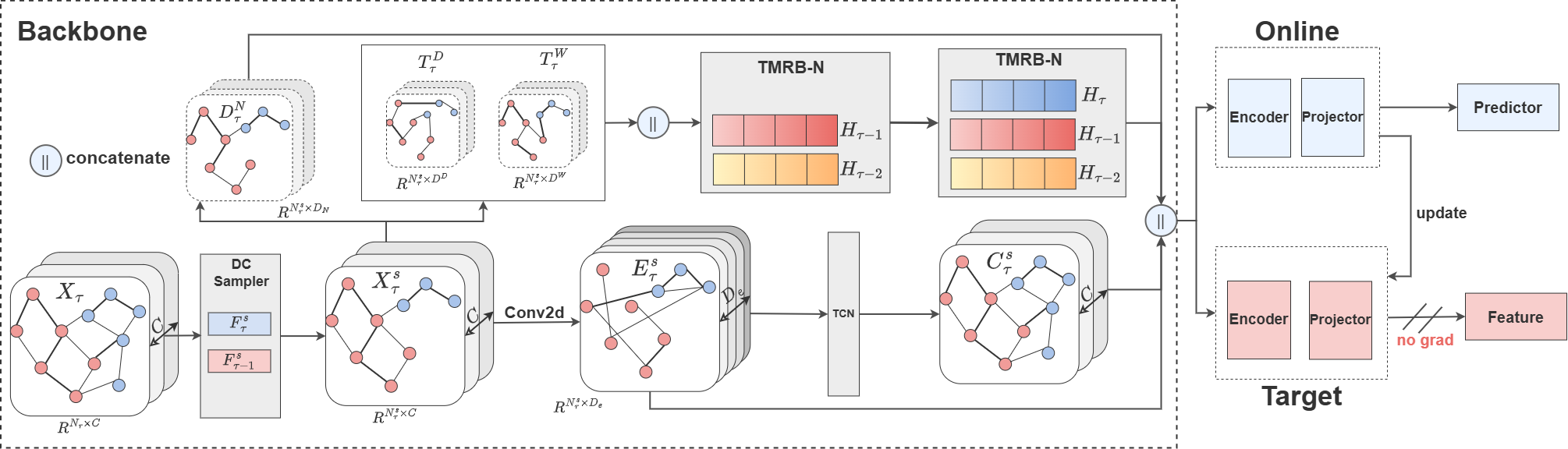}
    {\footnotesize Edges are shown only for node correspondence and optional update-set expansion; the predictor receives no adjacency matrix.\par}
    \caption{Overview of CoMemNet at period $\tau$. The Online branch is optimized by the current-period prediction loss, while the EMA Target branch provides a stop-gradient reference for drift scoring. The Drift Sampler selects drift-sensitive shared nodes, combines them with newly added nodes, and optionally expands the update set using sensor adjacency metadata. TMRB-N reuses compact temporal states to update $H_\tau$. The prediction backbone receives no adjacency matrix.}
    \label{fig:overview}
\end{figure*}

\subsection{Continual Learning}
Continual learning commonly relies on regularization, memory replay, or parameter isolation\cite{CL_1,CL_2,CL_3}. Regularization methods such as EWC\cite{CL_4} and LwF\cite{CL_5} constrain parameter or output changes, parameter-isolation methods allocate task-specific capacity\cite{CL_6,PackNet,CPG}, and replay methods retain representative samples or states\cite{CL_7,ER,iCaRL,DGR}. For evolving traffic prediction, TrafficStream\cite{Trafficstream}, PECPM\cite{PECPM}, TFMoE, and EAC\cite{expand} study continual adaptation under changing spatio-temporal patterns. Related dynamic sensor-network studies\cite{DGCLink,MCGCNet,ZhuTNSRE2025} further motivate nonstationary network modeling, but they are not direct traffic-forecasting baselines. Compared with these lines, CoMemNet emphasizes bounded shared-node selection, drift-aware update allocation, and explicit accounting of adaptation cost under period-evolving traffic sensors.

\section{Preliminaries}
\textbf{Expanding Network}: An expanding traffic network is represented as \( G = \{G_1, G_2, \dots, G_\tau\} \), where \( \tau \in \{1, 2, \dots, T\} \) represents the time period index (e.g., years). The network evolution process at time period \( \tau \) is represented as \( G_\tau = G_{\tau-1} + \Delta G_\tau \), where \( G_\tau = \{V_\tau, E_\tau, A_\tau\} \), where \( V_\tau \) is the set of nodes, satisfying \( |V_\tau| = N_\tau \), and \( N_\tau \) is the number of nodes at time \( \tau \), \( E_\tau \) is the set of edges, and the adjacency matrix \( A_\tau \in \mathbb{R}^{N_\tau \times N_\tau} \).

\textbf{Continual Prediction}: Given the streaming traffic signals $X=\{X_{(t-T_h+1:t)}^1,X_{(t-T_h+1:t)}^2,\dots,X_{(t-T_h+1:t)}^\tau\}$ and the expanding network $G$, the model $F$ predicts the future traffic flow $Y=\{Y_{(t+1:t+T_f)}^1,Y_{(t+1:t+T_f)}^2,\dots,Y_{(t+1:t+T_f)}^\tau\}$.

\section{Methodology}
As shown in Fig.~\ref{fig:overview}, CoMemNet contains a prediction backbone, an Online branch, an EMA Target branch, a Drift Sampler, and TMRB-N. For period $\tau$, the Drift Sampler applies a bounded shared-node selection rule and then forms the final update subset $V_\tau^s$ by adding newly added nodes and optional neighborhood expansion. Thus, $\rho$ controls the budget for drift-selected shared nodes, while the final update-set size can be larger after adding new nodes and expanded neighbors. The prediction backbone receives traffic and temporal/node embeddings but no adjacency matrix. Sensor adjacency is used only outside the predictor to construct the data and to expand selected nodes to a limited neighborhood when this option is enabled.

The Online branch is optimized for current-period prediction, whereas the Target branch is updated by exponential moving average and serves as a smoother reference for node ranking. The Online--Target discrepancy does not directly estimate the raw distributional distance between periods $\tau-1$ and $\tau$. Instead, it measures the disagreement between fast- and slow-updated representations under current-period inputs, and is used as a practical proxy for adaptation demand. No contrastive loss, positive/negative pair construction, or temperature term is used.
% \vspace{-0.2cm}

\subsection{Online-Target Branches}

The two branches provide representations with different update speeds. This makes their distributional discrepancy useful as a drift score, while the predictive objective remains unchanged. The ablations in Section~\ref{sec:revision_studies} show that the numerical benefit of the Target branch is modest; we therefore interpret it as a stabilizing reference rather than an indispensable source of accuracy.

\begin{figure}
    \centering
    \includegraphics[width=.99\linewidth]{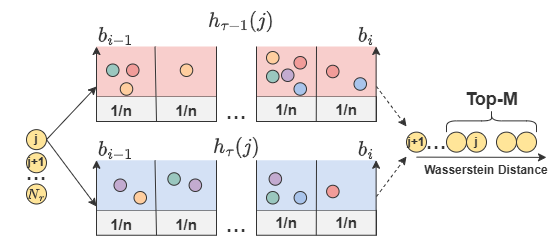}
    \caption{The Drift Sampler ranks shared nodes by node-wise Wasserstein-1 representation discrepancy, selects the top-$M$ shared nodes, and combines them with newly added nodes. Optional $K_{\mathrm{hop}}$-hop expansion can enlarge the final update subset.
}
    \label{fig:drift_sampler}
\end{figure}

In both the Online and Target branches, the Encoder is a Multi-Layer Perceptron (MLP) with \( L \) layers, used to extract features from the input data. The Projector is a $1\times1$ convolutional layer that performs point-wise channel projection without mixing information across nodes. The backbone transfers shared forecasting parameters across nodes and uses node-specific embeddings to represent sensor identity; it does not explicitly propagate information between adjacent nodes during prediction.

The Online branch is optimized on task $\tau$ using gradient descent:
\begin{equation}
    \theta_o \leftarrow \theta_o - \alpha \nabla_{\theta_o} L(\hat{Y}_{\tau}, Y_{\tau}, \theta_o)
\end{equation}
where $\alpha$ is the learning rate, $\nabla_{\theta_o} L$ denotes the gradient of the loss with respect to Online branch parameters, $\hat{Y}_{\tau}$ represents predicted traffic flow, and $Y_{\tau}$ is ground truth. This enables rapid convergence on current traffic patterns.

The Target branch updates via exponential moving average (EMA):
\begin{equation}
    \theta_t \leftarrow \beta \cdot \theta_t + (1 - \beta) \cdot \theta_o
\end{equation}
where $\beta \in [0, 1]$ controls the update rate. Higher $\beta$ (e.g., 0.99) retains a larger proportion of historical parameters, making updates slower and more stable. At each iteration, the Target branch preserves $\beta$ of its previous state while incorporating $(1-\beta)$ from the Online branch, acting as a low-pass filter that suppresses rapid fluctuations.

\begin{figure*}[t]
    \centering
    \includegraphics[width=.85\linewidth]{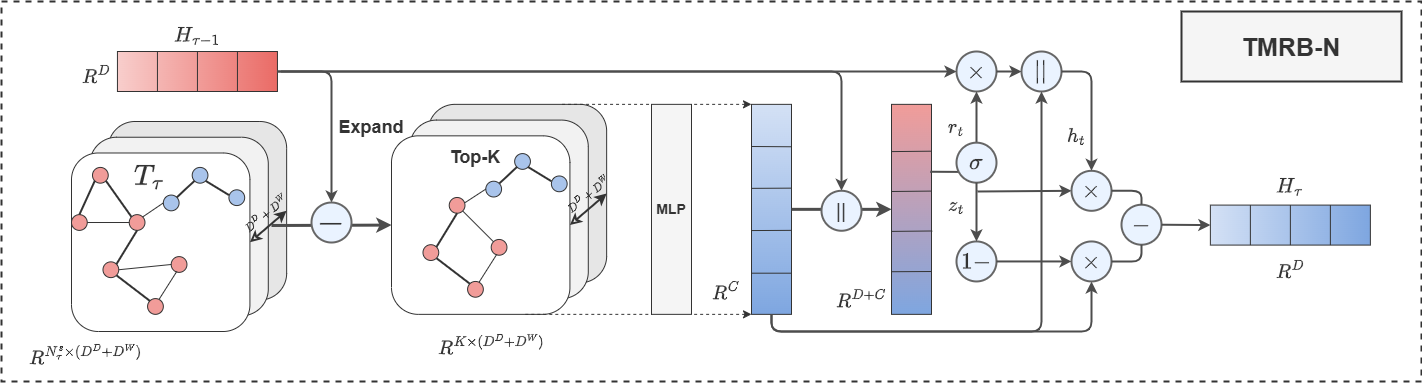}
    \caption{TMRB-N at period $\tau$. The module selects key temporal-state nodes, aggregates the current auxiliary state, and updates the compact latent state $H_\tau$, which is retained for temporal-state reuse in the next period.
}
    \label{fig:tmrb}
\end{figure*}
\subsection{Backbone}
For the selected node set $V_\tau^s$, the traffic input is encoded by the shared history encoder and point-wise projector, concatenated with node embeddings, temporal-convolution features, and the expanded TMRB-N state, and then mapped to $T_f$-step forecasts:
\begin{equation}
\begin{aligned}
E_\tau^s &= \operatorname{Conv}_{1\times1}(X_\tau^s),\\
Z_\tau^s &= E_\tau^s \parallel E_{\mathrm{node}}^s \parallel \operatorname{TCN}(E_\tau^s) \parallel \operatorname{Expand}(H_\tau,N_\tau^s),\\
\hat Y_\tau^s &= \operatorname{Projector}(\operatorname{Backbone}(Z_\tau^s)).
\end{aligned}
\end{equation}
The forecasting parameters are updated on $V_\tau^s$ and evaluated on the complete node set $V_\tau$. The node embedding is a shared learnable sensor-identity parameter broadcast to the current node set, so newly added nodes are adapted through current-period updates. No adjacency matrix is provided to this backbone.

\subsubsection{Drift Sampler}
The Drift Sampler selects newly added nodes and shared nodes with large Online--Target feature-distribution shifts. Its purpose is bounded drift-node selection for the shared nodes, followed by inclusion of newly added nodes and optional topology-assisted expansion. It is not a contrastive-learning objective, and the selected nodes are not assumed to be caused by topology changes.

Under our expanding-network protocol, $V_{\tau-1}\subseteq V_\tau$. The procedure, illustrated in Fig.~\ref{fig:drift_sampler}, selects all newly added nodes from $V_\tau\setminus V_{\tau-1}$ and a budgeted subset of shared nodes from $V_{\tau-1}\cap V_\tau$.

\textbf{Feature Extraction:} For every shared node $j\in V_{\tau-1}\cap V_\tau$, the Online and stop-gradient Target branches produce feature vectors $F_o^\tau(j)$ and $F_t^\tau(j)$. These vectors are flattened only along their feature coordinates to form two empirical one-dimensional distributions. This representation is a distribution-shift proxy and does not preserve the complete temporal ordering of a traffic profile.

\textbf{Shared Normalization and Discretization:} To avoid a distance induced by separate rescaling, both feature vectors of node $j$ use the same support:

\begin{equation}
\begin{aligned}
    \widetilde F_b^\tau(j)&=
    \frac{F_b^\tau(j)-a_j}{c_j-a_j+\varepsilon},\\
    a_j&=\min\{F_o^\tau(j),F_t^\tau(j)\},\\
    c_j&=\max\{F_o^\tau(j),F_t^\tau(j)\},
\end{aligned}
\label{equation2}
\end{equation}

where $b\in\{o,t\}$ and $\varepsilon$ prevents division by zero. The normalized coordinates are assigned to $B$ common ordered bins with support points $z_1<\cdots<z_B$, producing probability masses
\begin{equation}
    p_{b,j}(z_i)=\frac{1}{C_F}\sum_{c=1}^{C_F}
    \mathbf{1}\!\left[\widetilde F_{b,c}^\tau(j)\in \mathcal{B}_i\right],
    \label{equation3}
\end{equation}

where $\sum_i p_{b,j}(z_i)=1$.

\textbf{Wasserstein-1 Score Calculation:}
For one-dimensional ordered support, the discrete Wasserstein-1 distance equals the integral of the absolute difference between cumulative distribution functions. We compute the node-wise score
\begin{equation}
    d_j^\tau
    =\sum_{i=1}^{B-1}
    \left|P_{o,j}(z_i)-P_{t,j}(z_i)\right|
    \left(z_{i+1}-z_i\right),
    \label{equation4}
\end{equation}

where $P_{b,j}(z_i)=\sum_{r=1}^{i}p_{b,j}(z_r)$. Unlike the previous weighted bin-wise absolute difference, Eq.~\ref{equation4} explicitly includes cumulative mass and ground distance along the scalar support.

\textbf{Node Selection:}
Let $V_\tau^{\mathrm{new}}=V_\tau\setminus V_{\tau-1}$ and $V_\tau^{\mathrm{sh}}=V_{\tau-1}\cap V_\tau$. The $M$ shared nodes with the largest $d_j^\tau$ are combined with newly added nodes. A limited $K_{\mathrm{hop}}$-hop neighborhood can then be included using adjacency metadata:
\begin{equation}
\begin{gathered}
M = \lceil \rho |V_\tau^{\mathrm{sh}}|\rceil,\qquad
V_{\tau}^{\mathrm{drift}}=\operatorname{TopM}_{j\in V_\tau^{\mathrm{sh}}}d_j^\tau,\\
V_\tau^s=\operatorname{Expand}_{K_{\mathrm{hop}}}\!\left(
V_{\tau}^{\mathrm{drift}}\cup V_\tau^{\mathrm{new}}
\right).
\end{gathered}
\label{equation5}
\end{equation}

Here $\rho$ controls the budget for drift-selected shared nodes, not the final cardinality of $V_\tau^s$. Newly added nodes and optional $K_{\mathrm{hop}}$-hop expansion may increase the final update-set size. Attribution results show that $d_j^\tau$ is associated more strongly with traffic-statistical changes than with adjacency changes. We therefore call it a representation-drift proxy and make no topology-causality claim.
% \vspace{-0.1cm}

\subsubsection{TMRB-N}
TMRB-N performs compact latent-state replay, or temporal-state reuse, rather than raw-sample replay. It stores compressed temporal states from previous periods and reuses them in the current forward computation; raw historical sequences and labels are not revisited during incremental updating. This design provides temporal prior information and helps current-period adaptation, although it does not consistently reduce the standard forgetting metric reported in Sec.~\ref{sec:revision_studies}.

In the data processing section, we assign labels to each time point based on the daily and weekly time steps, creating daily and weekly time pools. In the input part of the model, for each specific time point, we generate daily embedding matrices \( T_\tau^D \in \mathbb{R}^{N_\tau^s \times D^D} \) and weekly embedding matrices \( T_\tau^W \in \mathbb{R}^{N_\tau^s \times D^W} \).

We concatenate the time features $T_\tau=(T_\tau^D\parallel T_\tau^W)\in\mathbb{R}^{N_\tau^s\times D_T}$, where $D_T=D^D+D^W$, and input them into TMRB-N (Fig.~\ref{fig:tmrb}). TMRB-N stores one compact vector $H_{\tau-1}\in\mathbb{R}^{D}$ from the previous period. In our implementation $D_T$ is projected to $D$ before state aggregation. Since $V_\tau^s$ changes across periods, the previous state is expanded to the current selected-node size only for computing node-wise differences.

To address these challenges, we use the following update mechanism to compute the time features:

\textbf{Feature Difference Calculation:}
First, the current temporal feature is projected to $D$ dimensions by $\bar T_\tau=T_\tau W_p$, where $W_p\in\mathbb{R}^{D_T\times D}$. The previous compact state is expanded to the selected-node size, and each node receives a scalar discrepancy score:
\begin{equation}
\begin{aligned}
    \Delta H_\tau(i,:)&=\left|\bar T_\tau(i,:)-\operatorname{expand}(H_{\tau-1},N_\tau^s)(i,:)\right|,\\
    q_{\tau,i}&=\left\|\Delta H_\tau(i,:)\right\|_2,
\end{aligned}
\end{equation}
where $H_{\tau-1}\in\mathbb{R}^{D}$ represents the compact temporal state from the previous period.

\textbf{Key Node Selection:}
By sorting the scalar scores $q_{\tau,i}$, TMRB-N selects the $K_{\mathrm{mem}}$ nodes with the largest temporal-state differences:
\begin{equation}
    V_\tau^k = \operatorname{TopK}_{i\in V_\tau^s}(q_{\tau,i},K_{\mathrm{mem}}).
    \label{eq:key_nodes}
\end{equation}

\textbf{Weighted Averaging of Current Time Features:}
For the selected key nodes $V_\tau^k$, compute the average current auxiliary state:
\begin{equation}
    H_\tau^a = \frac{1}{K_{\mathrm{mem}}} \sum_{i \in V_\tau^k} \bar T_\tau(i,:).
\end{equation}
The projection matrix $W_p\in\mathbb{R}^{D_T\times D}$ aligns the temporal embedding dimension with the compact memory dimension.

\textbf{Temporal Feature Update:}
To combine historical and current time features, we update the compact state using a gated mechanism:
\begin{equation}
    \begin{gathered}
        r_\tau = \sigma(W_r (H_{\tau-1} \parallel H_\tau^a)),\\
        z_\tau = \sigma(W_z (H_{\tau-1} \parallel H_\tau^a)),\\
        \widetilde H_\tau = \tanh(W_t (H_\tau^a \parallel (H_{\tau-1} \odot r_\tau))), \\
        H_\tau = z_\tau \odot \widetilde H_\tau + (1-z_\tau)\odot H_\tau^a .
    \end{gathered}
\end{equation}
where $W_r$, $W_t$, and $W_z$ denote learned weight matrices, $\sigma$ is the sigmoid activation, and $\odot$ is element-wise multiplication. The residual term is $H_\tau^a$ because the module stores an updated current-period auxiliary state after filtering it with the previous compact state. The updated $H_\tau$ is retained for period $\tau+1$.

\subsection{Loss Function and Evaluation Metrics}
The final model is optimized solely with Mean Absolute Error (MAE) on the node subset $V_\tau^s$ selected by the Drift Sampler. No contrastive or distillation loss is included:

\begin{equation}
    \resizebox{.98\linewidth}{!}{$
        \displaystyle
        \mathcal{L}_{\mathrm{MAE}} = \frac{1}{T_f |V_{\tau}^{s}| C} \sum_{t=1}^{T_f} \sum_{n\in V_{\tau}^{s}} \sum_{c=1}^{C} \left| \hat{Y}_{\tau}(t, n, c) - Y_{\tau}(t, n, c) \right|
    $}
\end{equation}

$\hat{Y}_\tau(t,n,c)$ and $Y_\tau(t,n,c)$ denote prediction and ground truth, respectively. Validation and test metrics are always computed over the entire available node set $V_\tau$, not only over selected nodes.

However, for the evaluation metrics on the validation and test sets, we use the entire node set \( V_{\tau} \):

\begin{equation}
\resizebox{.98\linewidth}{!}{$
    \displaystyle
    \begin{gathered}
        \text{MAE}(\hat{Y}_{\tau}, Y_{\tau}) = \frac{1}{T_f \times N_{\tau} \times C} \sum_{t=1}^{T_f} \sum_{n=1}^{V_{\tau}} \sum_{c=1}^{C} \left| \hat{Y}_{\tau}(t, n, c) - Y_{\tau}(t, n, c) \right| \\
        \text{RMSE}(\hat{Y}_{\tau}, Y_{\tau}) = \sqrt{\frac{1}{T_f \times N_{\tau} \times C} \sum_{t=1}^{T_f} \sum_{n=1}^{V_{\tau}} \sum_{c=1}^{C} \left( \hat{Y}_{\tau}(t, n, c) - Y_{\tau}(t, n, c) \right)^2} \\
        \text{MAPE}(\hat{Y}_{\tau}, Y_{\tau}) = \frac{1}{T_f \times N_{\tau} \times C} \sum_{t=1}^{T_f} \sum_{n=1}^{V_{\tau}} \sum_{c=1}^{C} \frac{\left| \hat{Y}_{\tau}(t, n, c) - Y_{\tau}(t, n, c) \right|}{Y_{\tau}(t, n, c)}
    \end{gathered}
$}
\end{equation}
\begin{table*}[t]
\centering
\caption{Comparison of annual average metrics at different temporal granularities. STID-retrained and EAC are run with their original code settings on the frozen CoMemNet splits. For STID on PEMSD4(L), MAPE is recomputed with $|Y|>1$ to avoid near-zero denominator instability.}
\label{tab:main_results}
\resizebox{.97\textwidth}{!}{
\begin{tabular}{c|c|ccccccccccc}
\toprule
  &
& \multirow{2}{*}{\textbf{Model}}& \multicolumn{3}{c}{\textbf{15 min}} & \multicolumn{3}{c}{\textbf{30 min}} & \multicolumn{3}{c}{\textbf{60 min}} & \textbf{Running Time} \\ 
\cmidrule(lr){4-6} \cmidrule(lr){7-9} \cmidrule(lr){10-12} \cmidrule{13-13}
&  & & MAE & RMSE & MAPE(\%) & MAE & RMSE & MAPE(\%) & MAE & RMSE & MAPE(\%) & Total/Avg. (s) \\ 
 \cmidrule{1-13}
 \multirow{14}{*}{\textbf{PEMSD3(S)}}& \multirow{7}{*}{\textbf{Retrained}}
 & Retrained-STModel & 12.70 & 20.57 & 16.88 & 13.82 & 22.61 & 18.40 & 16.10 & 26.54 & 24.96 & 179.56/1.39 \\
 & & Static-STModel & 14.57 & 22.47 & 30.69 & 15.45 & 24.34 & 31.11 & 17.78 & 28.57 & 32.60 & 39.71/1.10 \\
 & &Retrained-STKEC &12.62 & 20.43 & 18.24  &13.75 & 22.51 & 19.63  &16.02 & 26.42 & 22.72  &310.11/2.25\\
 & &Static-STKEC &14.27 &21.78 &34.88 &15.41 &23.90 &36.25 &17.68 &27.92 &39.38 &67.79/1.47 \\
 & &STID-retrained &11.88 &\underline{19.36} &16.67 &\underline{12.54} &\underline{20.57} &18.27 &\underline{13.79} &\textbf{22.65} &21.63 &--- \\
 & &PatchTST-retrained &13.10 &21.64 &18.53 &14.45 &24.00 &19.52 &17.27 &28.72 &24.28 &--- \\
 & &iTransformer-retrained &12.48 &20.80 &19.12 &13.33 &22.27 &20.28 &15.25 &25.42 &24.11 &--- \\
\cmidrule{2-13}
 &\multirow{7}{*}{\textbf{Continual}}
 & Expansible-STModel &14.93 & 24.70 & 18.99 & 17.36 & 28.60 & 20.61 & 21.31 & 36.91 & 24.45
 & 428.8 / 1.23 \\
 &  &Expansible-STKEC &12.64 &20.45 &17.68 &13.78 &22.51 &18.98 &16.07 &26.44 &22.23 &282.90/1.86 \\
 & & TrafficStream &12.75 & 20.78 & 17.43 & 13.92 & 22.80 & 18.75 & 16.27 & 26.80 & 22.90 &306.97/1.76 \\
 & &STKEC &12.73 &20.56 &19.03  &13.90 &22.69 &20.54 &16.19 &26.61 &24.09 &321.31/2.69 \\
 & &PECPM &\underline{11.81} &19.39 &\textbf{15.91} &13.41 &22.17 &\textbf{17.07} &14.76 &25.03 &\textbf{18.30}&--- \\
 & &TFMoE &12.48 &20.48 &16.72 &13.93 &23.00 &18.13 &17.20 &28.36 &22.17 &--- \\
& &EAC &18.67 &27.44 &41.73 &19.31 &28.65 &40.48 &21.37 &32.22 &42.46 &--- \\
& &\textbf{CoMemNet} &\textbf{11.48} &\textbf{18.89} &\underline{16.54} &\textbf{12.24} &\textbf{20.40} &\underline{17.31} &\textbf{13.57} &\underline{22.94} &\underline{18.80} &\textbf{103.66/0.30} \\
\midrule
 \multirow{11}{*}{\textbf{PEMSD4(L)}}& \multirow{5}{*}{\textbf{Retrained}}
 & Retrained-STModel &19.76 &31.73 &14.90 &21.82 &35.24 &16.44 &25.79 &41.38 &19.69
 & 403.60/2.60 \\
 & & Static-STModel & 
24.67 &36.69 &23.48 &26.59 &39.86 &25.72 &30.73 &46.27 &30.81
 & 59.00/1.84 \\
 & &Retrained-STKEC &20.40 &32.44 &14.67 &22.38 &35.86 &15.95 &26.19 &41.80 &18.74
  &470.26/3.59\\
 & &Static-STKEC &24.21 &35.80 &21.73 &26.07 &38.98 &23.01 &29.35 &44.41 &25.48 &94.87/2.43 \\
 & &STID-retrained &\underline{18.60} &\underline{30.99} &\textbf{12.71} &\underline{20.11} &\underline{33.63} &\underline{14.09} &\underline{22.81} &\underline{37.84} &\underline{16.76} &--- \\
\cmidrule{2-13}
 &\multirow{6}{*}{\textbf{Continual}}
 & Expansible-STModel &20.20 &32.16 &15.54 &22.25 &35.73 &16.98 &26.42 &42.26 &20.79
 & 352.14/1.88 \\
 &  &Expansible-STKEC &20.46 &32.45 &14.57 &22.44 &35.85 &15.94 &26.20 &41.66 &18.74
 &409.18/2.51 \\
 & & TrafficStream &23.76 &35.86 &21.60 &25.32 &38.88 &22.64 &29.33 &45.43 &25.99
 &494.59/2.58 \\
 & &STKEC &20.39 &32.54 &14.27 &22.42 &35.98 &15.60 &26.35 &42.04 &18.47
 &656.55/4.05 \\
& &EAC &84.33 &107.75 &130.36 &84.95 &108.35 &129.98 &85.97 &109.72 &129.85 &--- \\
& &\textbf{CoMemNet} &\textbf{17.98} &\textbf{30.44} &\underline{12.92} &\textbf{19.41} &\textbf{33.05} &\textbf{13.92} &\textbf{22.00} &\textbf{37.38} &\textbf{15.86}
&\textbf{228.60/0.65} \\
\midrule
 \multirow{11}{*}{\textbf{PEMSD8(M)}}& \multirow{5}{*}{\textbf{Retrained}}
 & Retrained-STModel &15.14 &24.22 &17.32 &16.66 &27.00 &19.79 &19.61 &32.05 &24.53   & 136.80/1.26 \\
 & & Static-STModel &19.54 &28.76 &34.04 &20.60 &31.27 &33.36 &23.66 &37.28 &34.67   & 39.51/1.16 \\
 & &Retrained-STKEC &15.66 &24.67 &17.78 &17.07 &27.27 &18.60 &19.73 &31.87 &20.24    &290.78/1.86\\
 & &Static-STKEC &17.00 &26.06 &16.84 &18.24 &28.43 &17.86 &20.76 &32.88 &19.80   &78.28/1.40 \\
 & &STID-retrained &\underline{14.27} &\underline{23.09} &16.82 &\underline{15.49} &\underline{25.14} &19.72 &\underline{17.65} &\underline{28.51} &25.28 &--- \\
\cmidrule{2-13}
 &\multirow{6}{*}{\textbf{Continual}}
 & Expansible-STModel &17.29 &26.84 &29.29 &19.06 &30.43 &30.43 &23.05 &37.72 &34.80
 & 151.08/1.18 \\
 &  &Expansible-STKEC &15.37 &24.35 &\underline{15.83} &16.80 &26.96 &\underline{17.15} &19.60 &31.68 &\underline{19.78}   &286.86/1.60 \\
 & & TrafficStream &15.88 &25.00 &26.59 &17.38 &28.10 &26.74 &20.65 &34.09 &30.01   &143.53/1.48 \\
 & &STKEC &15.34 &24.29 &16.70 &16.75 &26.90 &17.68 &19.44 &31.53 &20.04
 &249.96/1.81 \\
& &EAC &16.10 &24.81 &21.55 &17.22 &26.92 &22.64 &19.55 &30.91 &25.37 &--- \\
& &\textbf{CoMemNet} &\textbf{13.88} & \textbf{22.70} & \textbf{15.28}  
&\textbf{15.01} &\textbf{24.83} & \textbf{16.12}  
&\textbf{17.03} &\textbf{28.41} &\textbf{17.82}  
 &\textbf{92.03/0.23} \\
\bottomrule
\end{tabular}
}
\end{table*}

\begin{table}[t]
\centering
\caption{PEMSD3(S) 12-step performance of recent static current-period forecasters under the revision protocol.}
\label{tab:recent_static_pemsd3}
\scriptsize
\setlength{\tabcolsep}{3.0pt}
\resizebox{\columnwidth}{!}{%
\begin{tabular}{lclc}
\toprule
Method & Batch size & Training setting & 12-step MAE/RMSE\\
\midrule
CoMemNet & 128 & 100-epoch cap, patience 50 & \textbf{13.42}/\textbf{22.72}\\
STID-current & 128 & 100-epoch cap, patience 20 & 14.20/23.32\\
DLinear-style current & 128 & 100-epoch cap, patience 20 & 22.85/37.44\\
PatchTST-Lite current & 32 & 100-epoch cap, patience 20 & 17.27/28.72\\
iTransformer-Lite current & 16 & 100-epoch cap, patience 20 & 15.25/25.42\\
\bottomrule
\end{tabular}}
\end{table}

\begin{table}[t]
\centering
\caption{Training hyperparameters for different datasets.}
\label{tab:hyperparameters}
    \resizebox{.48\textwidth}{!}{
\begin{tabular}{ccccccc}
\toprule[1.5pt]
Datasets & Batch Size & Lr  & Epoch & Momentum & Lr Decay & $\rho$ \\ \midrule
PEMSD3(S)& 128 & 0.01 & 50& 0.99& 0.5& 0.05 \\
PEMSD4(L) & 128& 0.01& 50& 0.99& 0.5 & 0.03 \\
PEMSD8(M) & 128& 0.01& 60& 0.99 & 0.5 & 0.05\\
\bottomrule[1.5pt]
\end{tabular}
}
\end{table}

\begin{table}[t]
    \centering
    \caption{Statistics of PEMS datasets}
    \label{tab:datasets}
    \resizebox{.48\textwidth}{!}{
    \begin{tabular}{ccccccccc}
    \toprule[1.5pt]
    \multirow{3}{*}{PEMSD3(S)} &Year &2011 &2012 &2013 &2014 &2015 &2016 &2017 \\
    &\#Node & 655& 715& 786& 822& 834& 850& 871\\
    &\#Edges & 1577& 1929& 2316& 2536& 2594 &2691&2788\\ 
    \midrule
    \multirow{3}{*}{PEMSD4(L)} &Year &2009 &2010 &2011 &2012 &2013 &2014 &2015 \\
    & \#Nodes& 1118 & 1687 & 1742& 1785& 1799& 2252 & 2406\\ 
    & \#Edges& 2437& 4746& 5095& 5330& 5380& 8401&9773\\
    \midrule
     \multirow{3}{*}{PEMSD8(M)} &Year &2012 &2013 &2014 &2015 &2016 &2017 &2018 \\
    & \#Nodes& 216& 248 & 267& 274& 297& 318 & 320\\ 
    & \#Edges& 488& 619& 744& 788& 901& 1076& 1089\\ 
    \bottomrule[1.5pt]
    \end{tabular}}
\end{table}

\section{Experiment}
\subsection{Datasets}
As shown in Table~\ref{tab:datasets}, we conduct experiments on three real-world traffic datasets: PEMSD3(S), PEMSD4(L), and PEMSD8(M). PEMSD3(S) follows the public PEMSD3-stream evolving traffic benchmark released by prior work, rather than an arbitrary split constructed by us. PEMSD4(L) and PEMSD8(M) are built from raw CalTrans PeMS traffic detector records to extend continual-learning evaluation to different regions, network scales, and sensor-evolution gradients. All three datasets use real sensor observations sampled at 5-minute intervals; no synthetic nodes or synthetic traffic flows are introduced. Our task is to predict traffic flow for the next 60 minutes (12 time steps) from the past 60 minutes (12 time steps).

\subsection{Dataset Processing}

For PEMSD4(L) and PEMSD8(M), processing starts from CalTrans PeMS 5-minute detector records. We organize records by year, align cross-year nodes by sensor ID, and select sensors using three criteria: missing rate below 10\%, State Post-Mile below 100, and persistence from year $\tau$ to $\tau+\Delta\tau$ ($\Delta\tau=1$). Minor gaps are filled by linear interpolation. The persistence criterion is used only for offline benchmark curation to remove transient sensors; it is shared by all methods and exposes no future traffic values, labels, validation results, or node-selection signals during continual training.

The predefined adjacency matrix for year $\tau$ is constructed from sensor metadata as follows:
\[
A_{\tau}[m,n] =
\begin{cases}
\exp\left(-\frac{d_{mn}^2}{\delta^2}\right),
& \substack{m\neq n,\ r_m=r_n,\ q_m=q_n,\\ d_{mn}<\epsilon},\\
0, & \text{otherwise},
\end{cases}
\]
where $d_{mn}=|\mathrm{postmile}_m-\mathrm{postmile}_n|$, $r_m$ denotes the route/highway identifier, and $q_m$ denotes the travel direction. State Post-Mile is meaningful only for sensors on comparable road segments; therefore, two sensors are connected only when their route and direction match and their Post-Mile distance is below $\epsilon$. The parameter $\delta$ (set to 100) controls the Gaussian spatial influence decay, and $\epsilon$ (set to 1) defines the sparsity threshold. This produces a sparse metadata adjacency used for data construction and optional update-set expansion, not as an input to the prediction backbone.

\subsection{Experimental Setting}
The experiments were conducted on systems equipped with an NVIDIA GeForce RTX 4090 GPU and a vGPU-32GB/16-vCPU Intel(R) Xeon(R) environment. Each year's data was split along the temporal dimension into training, validation, and testing sets with a ratio of 6:2:2, ensuring temporal order was preserved to avoid data leakage. The training process utilized the AdamW optimizer, with a learning rate scheduling strategy to dynamically adjust the learning rate. A momentum mechanism was employed to accelerate the convergence of the target branch. To prevent overfitting, an early stopping mechanism was applied with a patience of 10 epochs. The hyperparameters related to training are shown in Table~\ref{tab:hyperparameters}.

\subsubsection{Continual-Learning Protocol}
Table~\ref{tab:protocol} summarizes the information available to each update protocol. For every period $\tau$, training, validation, and test windows are temporally ordered. Model selection uses only the validation split of period $\tau$; no future-period sequence, label, or validation result is used. CoMemNet may retain the previous model, the EMA Target parameters, compact TMRB-N state, and sensor metadata required to align nodes. It does not jointly load all historical raw training sequences during an incremental update. Current-only CoMemNet-backbone starts from random initialization and uses only period-$\tau$ training data, whereas All-history CoMemNet-backbone starts from random initialization and reads all training periods up to $\tau$.

\begin{figure*}[t]
    \centering
    \includegraphics[width=.98\textwidth]{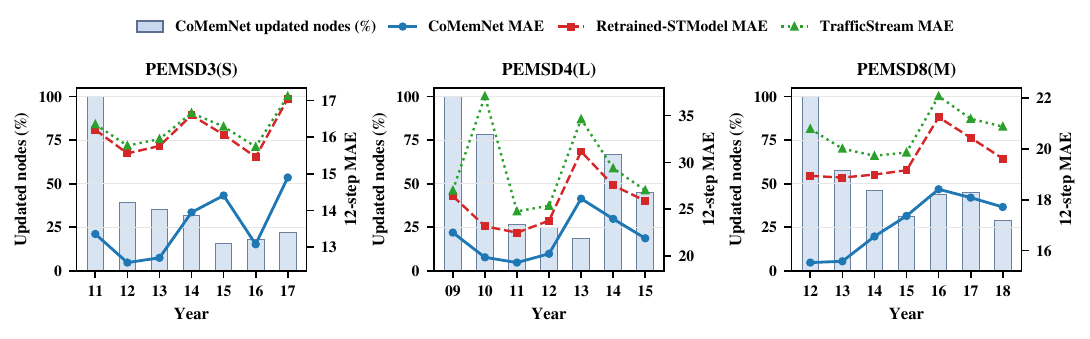}
    \caption{Period-wise update scale and 12-step MAE across the three evolving PeMS datasets. Bars report the percentage of nodes updated by CoMemNet relative to the current-period node set; lines report 12-step MAE for CoMemNet and representative retrained/continual baselines. The figure replaces the long node-count table and highlights the accuracy--efficiency trade-off across periods.}
    \label{fig:update_efficiency}
\end{figure*}

\begin{table}[t]
\centering
\caption{Information available when learning period $\tau$.}
\label{tab:protocol}
\resizebox{.48\textwidth}{!}{
\begin{tabular}{lccc}
\toprule
Information & CoMemNet & Current-only backbone & All-history backbone\\
\midrule
Current training/labels & Yes & Yes & Yes\\
Previous model/EMA state & Yes & No & No\\
Compact TMRB-N state & Yes & No & No\\
Historical raw training data & No & No & Yes\\
Current validation split & Yes & Yes & Yes\\
Future-period training signals & No & No & No\\
Sensor metadata/adjacency & Update set only & Preprocessing & Preprocessing\\
\bottomrule
\end{tabular}}
\end{table}

\subsection{Baseline Methods}

\textbf{Alternative Backbone Models:} STModel\cite{Trafficstream} and STKEC\cite{STECK} are spatio-temporal backbone models evaluated under three training paradigms to establish performance bounds:

\begin{itemize}
    \item \textbf{Retrained-STModel/Retrained-STKEC}: The model is retrained from scratch using all historical data available up to the current period. Its cumulative cost grows with both history length and update frequency; we do not assume that a single annual retraining run is universally infeasible.
    
    \item \textbf{Static-STModel/Static-STKEC}: The model trained only on the first year's data is applied to all subsequent years without updates. This represents a lower bound, illustrating severe performance degradation when the model cannot adapt to evolving patterns and topology changes.
    
    \item \textbf{Expansible-STModel/Expansible-STKEC}: The model incrementally trains only on newly added nodes $(V_\tau \setminus V_{\tau-1})$ and their two-hop neighbors, using the previous year's model as initialization. This is an efficient adaptation reference without explicit memory consolidation.
\end{itemize}

\noindent \textbf{Continual Learning Methods:} We also compare against state-of-the-art continual learning methods specifically designed for evolving traffic networks:

\begin{itemize}
    \item \textbf{TrafficStream}\cite{Trafficstream}: An online continual learning framework that integrates new patterns through dynamic pattern fusion and sliding window-based knowledge consolidation.
    
    \item \textbf{STKEC}\cite{STECK}: A knowledge expansion and consolidation framework using impact-based strategies to preserve critical knowledge while accommodating new patterns.
    
    \item \textbf{PECPM}\cite{PECPM}: A pattern bank-based framework that stores and retrieves representative spatio-temporal patterns for efficient knowledge reuse.
    
    \item \textbf{TFMoE}: A Mixture-of-Experts approach that assigns specialized models to homogeneous traffic groups for continual prediction.
    
    \item \textbf{EAC}\cite{expand}: A prompt-based method that learns task-specific parameters while keeping the backbone frozen, balancing adaptation and preservation.
\end{itemize}

\noindent \textbf{Static Retraining Baselines:} Static retraining baselines include STID-retrained\cite{STID}, PatchTST-retrained\cite{PatchTST}, and iTransformer-retrained\cite{iTransformer}. For each period, these models are initialized from scratch, trained on the frozen current-period training split, selected using the same-period validation split, and evaluated on the same-period test split. STID is a lightweight MLP-based predictor with spatial and temporal identity embeddings. PatchTST encodes temporal patches for each node with a Transformer, while iTransformer treats nodes as tokens and computes node-wise attention.

\subsection{Experiment Results}
In this section, we compare CoMemNet with baseline models in terms of effectiveness and efficiency.
\subsubsection{Effectiveness}
Table \ref{tab:main_results} compares the annual average MAE, RMSE, and MAPE performance of different baseline models across three temporal granularities (15 minutes, 30 minutes, and 60 minutes). The results show that the CoMemNet model significantly outperforms other baseline models in most metrics.

Specifically, the Static alternative cannot adapt to later-period distributions. Retrained alternatives provide strong accuracy references but repeatedly process increasing amounts of data. The Expansible alternative updates newly added nodes without explicit memory consolidation. CoMemNet achieves a favorable current-period accuracy and cumulative-cost trade-off; the more detailed continual-learning metrics below show that this advantage should not be interpreted as uniformly lower forgetting.

\subsubsection{Efficiency}
\paragraph{Update scale of CoMemNet.}
Fig.~\ref{fig:update_efficiency} visualizes the period-wise update scale and 12-step MAE. The Drift Sampler restricts shared-node selection while maintaining competitive prediction accuracy. Together with the running time in Table~\ref{tab:main_results}, Fig.~\ref{fig:update_efficiency} shows the practical efficiency--accuracy trade-off within CoMemNet's own update process. For example, on PEMSD3(S), which expands from 655 to 871 nodes, CoMemNet updates only 130--192 nodes in 2015--2017 while preserving strong prediction accuracy.

\paragraph{Relative advantage under larger node scales.}
We further evaluate 25\%, 50\%, 75\%, and 100\% node-scale variants of PEMSD4(L), constructed with fixed node budgets while preserving the original yearly order and real traffic records.

\begin{figure}[t]
    \centering
    \includegraphics[width=.98\linewidth]{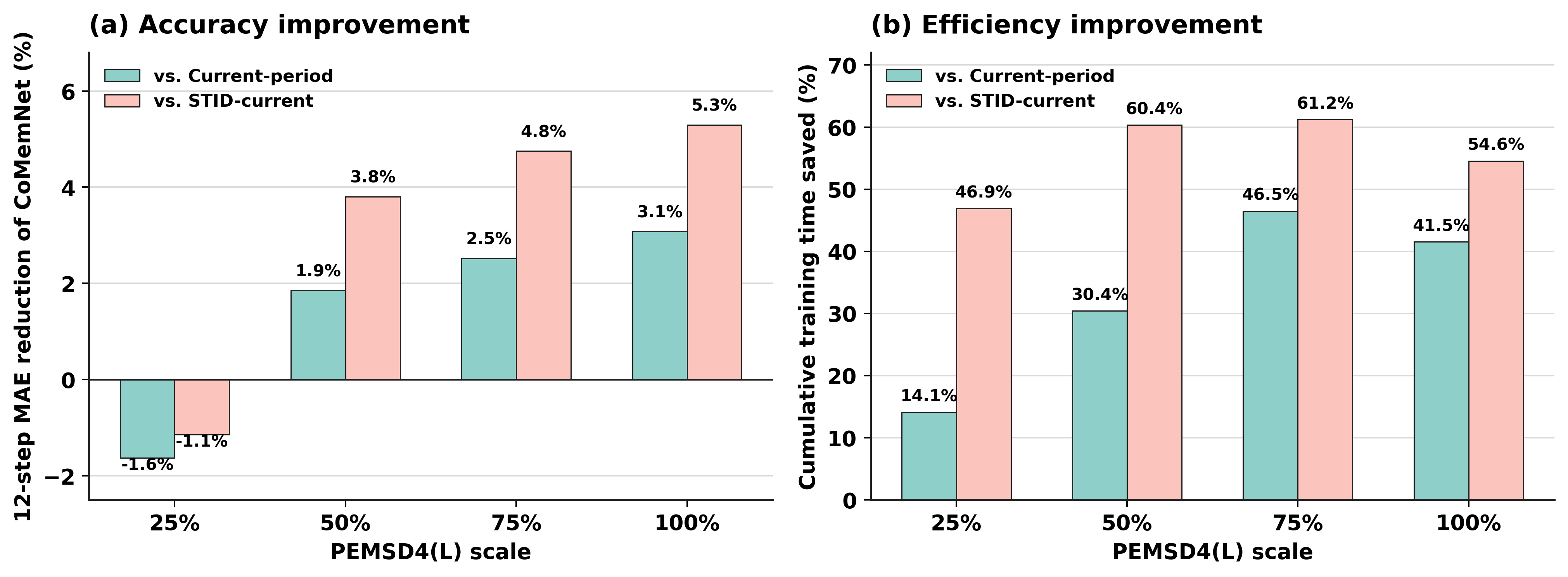}
    \caption{Relative percentage advantages of CoMemNet on PEMSD4(L) scale variants.}
    \label{fig:round2_scaling_advantage}
\end{figure}

Fig.~\ref{fig:round2_scaling_advantage} shows the advantage over current-period retraining, including lightweight static retraining baselines such as STID-current: CoMemNet updates only new and drift-sensitive nodes rather than retraining all current nodes, so the relative training-time saving becomes more pronounced as the node scale increases, while the 12-step MAE advantage also becomes clearer.

\subsection{Ablation Study}
\subsubsection{Impact of Drift Sampling}

The Drift Sampler combines newly added nodes $V_\tau\setminus V_{\tau-1}$ with shared nodes ranked by the node-wise Wasserstein-1 score.

\begin{itemize}
    \item \textbf{Static CoMemNet}: Utilizes the CoMemNet model trained on the first year to directly infer the remaining years.
    \item \textbf{Retrained CoMemNet}: Trains the model each year using all nodes.
    \item \textbf{w/o Increase}: Excludes the newly added nodes \((V_\tau \setminus V_{\tau-1})\).
    \item \textbf{w/o Drifted Shared Nodes}: Excludes the shared drift-sensitive nodes selected for updating and updates only the newly added nodes.
\end{itemize}

\begin{table}[t]
    \centering
       \caption{Ablation Study on Drift Sampling: Annual Average 12-Step Mean Performance Metrics}
    \resizebox{.45\textwidth}{!}{
    \begin{tabular}{c|cccc}
    \toprule[1.5pt]
         & Variants &MAE &RMSE &MAPE(\%)\\ 
         \midrule
         \multirow{5}{*}{\textbf{PEMSD3(S)}}
         &Static CoMemNet & 35.81&51.23 &33.11 \\
         &Retrained CoMemNet & \textbf{13.00}&\textbf{21.81}&\textbf{18.59} \\
         &w/o Increase &14.17 & 24.09&21.61 \\
         &w/o Drifted Shared Nodes & 14.65&24.67 &19.39 \\
         &\textbf{CoMemNet} &\underline{13.57} &\underline{22.94} &\underline{18.80} \\
         \midrule
                  \multirow{5}{*}{\textbf{PEMSD4(L)}}
         &Static CoMemNet &35.29 &52.75 &23.55 \\
         &Retrained CoMemNet & \textbf{21.62}& \textbf{36.84}&\textbf{15.37} \\
         &w/o Increase & 22.50& 38.07&15.90 \\
         &w/o Drifted Shared Nodes & 22.33&37.71 &16.20 \\
         &\textbf{CoMemNet} & \underline{22.00}& \underline{37.38}&\underline{15.86} \\
         \midrule
                  \multirow{5}{*}{\textbf{PEMSD8(M)}}
         &Static CoMemNet &34.83 & 50.10& 25.56\\
         &Retrained CoMemNet & \textbf{15.82}& \textbf{26.37}&\underline{18.50} \\
         &w/o Increase & 17.48&29.22 &19.86 \\
         &w/o Drifted Shared Nodes & 18.47&30.27 &20.00 \\
         &\textbf{CoMemNet} &\underline{17.03} &\underline{28.41} &\textbf{17.82} \\
         \bottomrule[1.5pt]
    \end{tabular}
    }

    \label{tab:drift_ablation}
\end{table}

Table~\ref{tab:drift_ablation} shows that Static CoMemNet fails to adapt to later distributions, whereas retraining gives a strong accuracy reference at higher cumulative cost. Removing newly added nodes or drifted shared nodes degrades current-period prediction. These results support the role of bounded shared-node updating in continual adaptation, but they do not by themselves establish lower standard forgetting; that question is evaluated directly in Table~\ref{tab:continual_controls}.

\subsubsection{Impact of TMRB-N}

In Table~\ref{tab:tmrb_ablation}, we conducted an ablation study on the TMRB-N module and its two key components: critical node selection and temporal feature update.
\begin{itemize}
\item \textbf{w/o TMRB-N}: Removes the TMRB-N module.
\item \textbf{w/o Select}: Replaces the feature difference-based critical node selection method in TMRB-N with random selection.
\item \textbf{w/o Update}: Removes the temporal feature update component from the TMRB-N module.
\end{itemize}

Removing the TMRB-N module across all datasets degrades performance, supporting the value of compact temporal-state reuse. Replacing the key-node selection method with random selection or removing the temporal update also reduces performance in most metrics. The complete model achieves the best overall performance, supporting the contribution of both components without implying that TMRB-N uniformly reduces the standard forgetting metric.
\begin{table}[t]
    \centering
       \caption{Ablation Study on the Impact of TMRB-N: Annual Average 12-Step Mean Performance Metrics}
    \resizebox{.43\textwidth}{!}{
    \begin{tabular}{c|cccc}
    \toprule[1.5pt]
         & Variants &MAE &RMSE &MAPE(\%)\\ 
         \midrule
         \multirow{4}{*}{\textbf{PEMSD3(S)}}
         &w/o TMRB-N & 14.94&25.38 &19.76 \\
         &w/o Select & 13.82&23.26&19.33 \\
         &w/o Update &13.63 & 23.09&\textbf{18.51} \\
         &\textbf{CoMemNet} &\textbf{13.57} &\textbf{22.94} &18.80 \\
         \midrule
                  \multirow{4}{*}{\textbf{PEMSD4(L)}}
         &w/o TMRB-N &23.76 &39.59 &17.29 \\
         &w/o Select & 22.14& 37.56&\textbf{15.85} \\
         &w/o Update & 22.15& 37.40&16.66 \\
         &\textbf{CoMemNet} & \textbf{22.00}& \textbf{37.38}&15.86 \\
         \midrule
                  \multirow{4}{*}{\textbf{PEMSD8(M)}}
         &w/o TMRB-N &18.36 & 30.61& 20.06\\
         &w/o Select & 17.33& 28.82&18.96 \\
         &w/o Update & 17.10&28.59 &18.29 \\
         &\textbf{CoMemNet} &\textbf{17.03} &\textbf{28.41} &\textbf{17.82} \\
         \bottomrule[1.5pt]
    \end{tabular}
    }

    \label{tab:tmrb_ablation}
\end{table}

\subsection{Parameter Sensitivity Analysis}

\subsubsection{Sensitivity Analysis of Hyperparameter $\rho$}

In Eq.~\ref{equation5}, $\rho$ controls the number of drift-selected shared nodes, not the final expanded update-set size. The final set can also include newly added nodes and optional $K_{\mathrm{hop}}$-hop neighbors. As shown in Table~\ref{tab:rho_sensitivity}, increasing $\rho$ usually improves current-period prediction but also increases computation. Some years are not monotonic, indicating that the budget is a deployment-dependent stability--plasticity hyperparameter rather than a universal constant.

\begin{table*}[t]
    \centering
    \caption{Comparison of Number of Nodes Trained Annually and 12-Step Average MAE Performance Across Different $\rho$ Variants}
    \resizebox{.98\textwidth}{!}{
    \begin{tabular}{c|ccccccccccccccc}
    \toprule[1.5pt]
    \multirow{8}{*}{\textbf{PEMSD3(S)}}&\multirow{2}{*}{\textbf{Variants}}&\multicolumn{2}{c}{\textbf{2011}}&\multicolumn{2}{c}{\textbf{2012}}&\multicolumn{2}{c}{\textbf{2013}} &\multicolumn{2}{c}{\textbf{2014}} &\multicolumn{2}{c}{\textbf{2015}} &\multicolumn{2}{c}{\textbf{2016}} &\multicolumn{2}{c}{\textbf{2017}} \\
    \cmidrule(lr){3-4}\cmidrule(lr){5-6}\cmidrule(lr){7-8}\cmidrule(lr){9-10}\cmidrule(lr){11-12}\cmidrule(lr){13-14} \cmidrule(lr){15-16}
    & &\textbf{Node}&\textbf{MAE}&\textbf{Node}&\textbf{MAE} &\textbf{Node}&\textbf{MAE}&\textbf{Node}&\textbf{MAE}&\textbf{Node}&\textbf{MAE}&\textbf{Node}&\textbf{MAE}&\textbf{Node}&\textbf{MAE} \\
    \cmidrule{2-16}
    &$\rho$=0&655&13.36 &182 &12.76 &146 &12.94 &151 &14.07&65 &16.50&31 &14.13&60  &18.78\\
    &$\rho$=0.03&655&13.36 &246&12.68 &212&12.74 &210 &14.05 &92&14.78 &96&13.25 &130 &14.89\\
    &$\rho$=0.05 (CoMemNet)  &655&13.36 &279 &12.58&275&12.71 &263 &13.95&130 &14.41&153 &13.08&192&14.90\\
    & $\rho$=0.08  &655 &13.36&328&12.55  &322 &12.61 &351&13.97 &191&14.28 &229&13.05 &263 &14.58\\
     & $\rho$=0.1  &655 &13.36 &360&12.48  &396 &12.62 &379&13.64 &245&14.11 &243&12.76 &338 &14.44\\
     & $\rho$=0.15  &655 &13.36&407&12.52  &410 &12.68 &445&13.59 &345&14.01 &356&12.59 &406 &14.26\\
    \midrule

        \multirow{8}{*}{\textbf{PEMSD4(L)}}&\multirow{2}{*}{\textbf{Variants}}&\multicolumn{2}{c}{\textbf{2009}}&\multicolumn{2}{c}{\textbf{2010}}&\multicolumn{2}{c}{\textbf{2011}} &\multicolumn{2}{c}{\textbf{2012}} &\multicolumn{2}{c}{\textbf{2013}} &\multicolumn{2}{c}{\textbf{2014}} &\multicolumn{2}{c}{\textbf{2015}} \\
    \cmidrule(lr){3-4}\cmidrule(lr){5-6}\cmidrule(lr){7-8}\cmidrule(lr){9-10}\cmidrule(lr){11-12}\cmidrule(lr){13-14} \cmidrule(lr){15-16}

    & &\textbf{Node}&\textbf{MAE}&\textbf{Node}&\textbf{MAE} &\textbf{Node}&\textbf{MAE}&\textbf{Node}&\textbf{MAE}&\textbf{Node}&\textbf{MAE}&\textbf{Node}&\textbf{MAE}&\textbf{Node}&\textbf{MAE} \\
    \cmidrule{2-16}
    
        &$\rho$=0&1118&22.33 &1293 &19.97 &311 &19.65&287 &20.67&109 &27.30&1459 &24.14&912  &22.11\\
    &$\rho$=0.03 (CoMemNet) &1118&22.52 &1322 &19.88&459&19.32 &447 &20.26&332 &26.14&1510 &23.98&1089&21.91 \\
    &$\rho$=0.05  &1118&22.52 &1350&19.90 &528&19.18&569 &20.08&467 &26.10&1534&23.96 &1153 &21.90\\
    & $\rho$=0.08  &1118&22.52&1385&19.88  &624 &19.05 &694&19.94 &635&25.96 &1577&23.85 &1262 &21.87\\
     & $\rho$=0.1  &1118&22.52&1403&19.85  &703 &19.06 &765&19.96 &683&25.85 &1609&23.91 &1309 &21.96\\
     & $\rho$=0.15  &1118&22.52 &1429&19.78  &845 &18.86 &930&19.81 &807&25.80 &1686&23.77 &1494 &21.75\\
    \midrule

        \multirow{8}{*}{\textbf{PEMSD8(M)}}&\multirow{2}{*}{\textbf{Variants}}&\multicolumn{2}{c}{\textbf{2012}}&\multicolumn{2}{c}{\textbf{2013}}&\multicolumn{2}{c}{\textbf{2014}} &\multicolumn{2}{c}{\textbf{2015}} &\multicolumn{2}{c}{\textbf{2016}} &\multicolumn{2}{c}{\textbf{2017}} &\multicolumn{2}{c}{\textbf{2018}} \\
    \cmidrule(lr){3-4}\cmidrule(lr){5-6}\cmidrule(lr){7-8}\cmidrule(lr){9-10}\cmidrule(lr){11-12}\cmidrule(lr){13-14} \cmidrule(lr){15-16}

    & &\textbf{Node}&\textbf{MAE}&\textbf{Node}&\textbf{MAE} &\textbf{Node}&\textbf{MAE}&\textbf{Node}&\textbf{MAE}&\textbf{Node}&\textbf{MAE}&\textbf{Node}&\textbf{MAE}&\textbf{Node}&\textbf{MAE} \\
    \cmidrule{2-16}
    
   &$\rho$=0&216&15.52 &119&15.64 &91 &17.53&43 &23.24&78 &18.98&116&18.57 &16 &19.82\\
    &$\rho$=0.05 (CoMemNet)&216&15.52 &143 &15.57&123&16.55 &85&17.36&131&18.41&143 &18.08&92&17.71 \\
    &$\rho$=0.1  &216&15.52 &151&15.41 &139 &15.86&124 &17.06&166 &18.05&184&18.19 &134 &17.22\\
    & $\rho$=0.15 &216&15.52&170&15.23  &162 &15.98 &156&16.18 &184&17.76 &192&17.77 &186 &17.02\\
     & $\rho$=0.2  &216&15.52&183&15.23  &180 &15.44 &173&15.76 &200&17.61 &208&17.70 &213 &16.77\\
     & $\rho$=0.3  &216&15.52&205&15.06  &215 &15.42 &196&15.64 &232&17.63 &261&17.48 &252 &16.60\\
    \bottomrule[1.5pt]
    \end{tabular}
    }
    \label{tab:rho_sensitivity}
\end{table*}

\subsubsection{Sensitivity Analysis of Hyperparameter $K_{\mathrm{mem}}$}

The effectiveness of the key-node selection method in Eq.~\ref{eq:key_nodes} was validated in Section 5.6. Here, $K_{\mathrm{mem}}$ denotes the number of key nodes used for TMRB-N temporal-state aggregation; it is distinct from the neighborhood expansion order $K_{\mathrm{hop}}$ in Eq.~\ref{equation5}.

\begin{figure}
    \centering
    \includegraphics[width=0.95\linewidth]{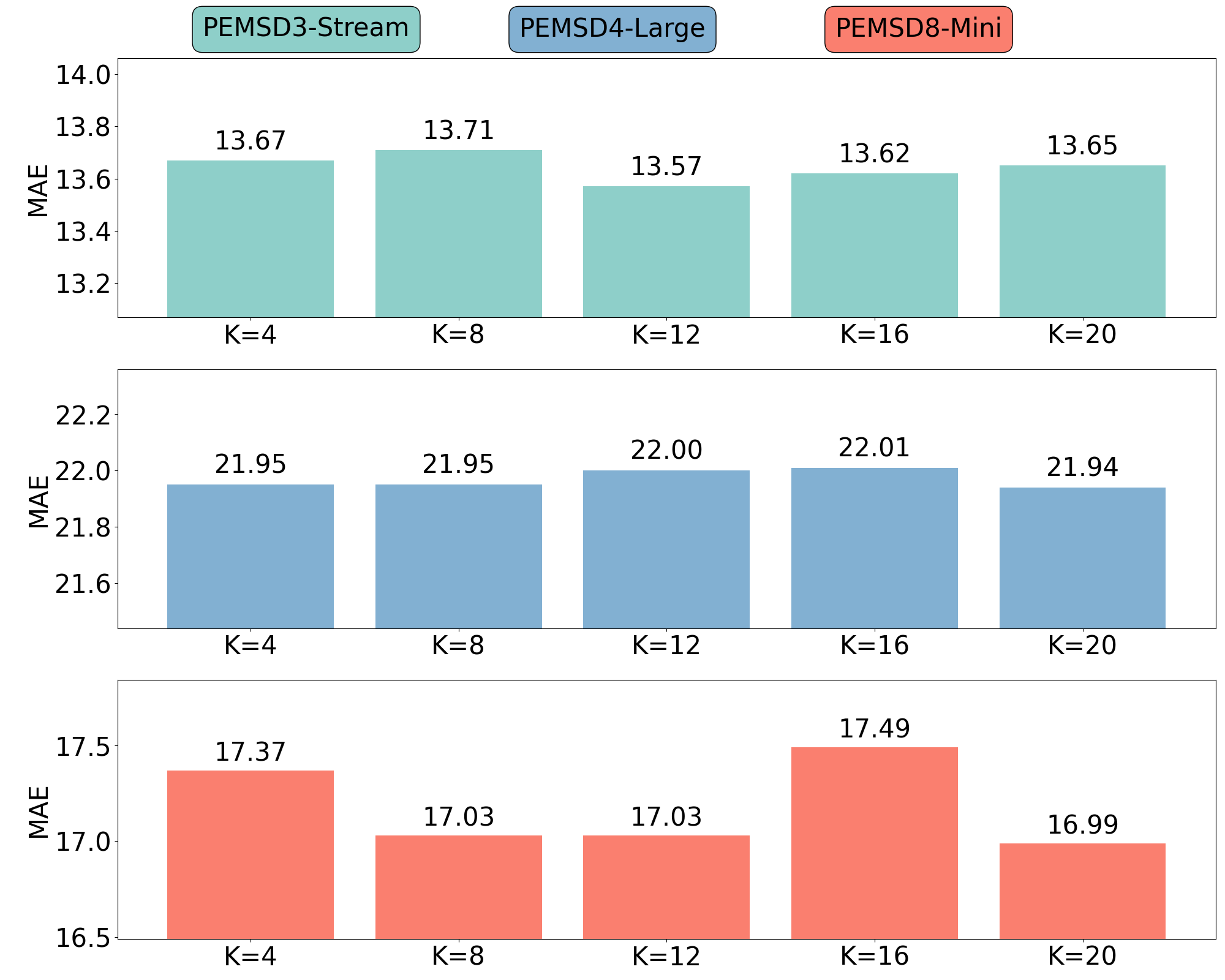}
    \caption{Sensitivity Analysis of Hyperparameter $K_{\mathrm{mem}}$: Annual Average 12-Step MAE}
    \label{fig:k_mem_sensitivity}
    \vspace{-0.8cm}
\end{figure}

Fig.~\ref{fig:k_mem_sensitivity} shows that $K_{\mathrm{mem}}=12$ gives a low or near-low MAE in the tested configurations. This value is an empirical setting rather than a scale-invariant property of traffic networks. The neighborhood controls in Fig.~\ref{fig:sampler_controls} further show only modest differences among selected-only and 1/2/3-hop expansion. We therefore treat both $\rho$ and $K_{\mathrm{hop}}$ as deployment-dependent update-policy hyperparameters, while $K_{\mathrm{mem}}$ controls the compact temporal-memory aggregation.

\section{Expanded Controlled Evaluation}
\label{sec:revision_studies}
\subsection{Drift-Score Visualization}
\begin{figure*}[t]
  \centering
  \setlength{\abovecaptionskip}{2pt}
  \setlength{\belowcaptionskip}{-2pt}
  \begin{subfigure}{0.315\textwidth}
    \centering
    \includegraphics[width=\linewidth]{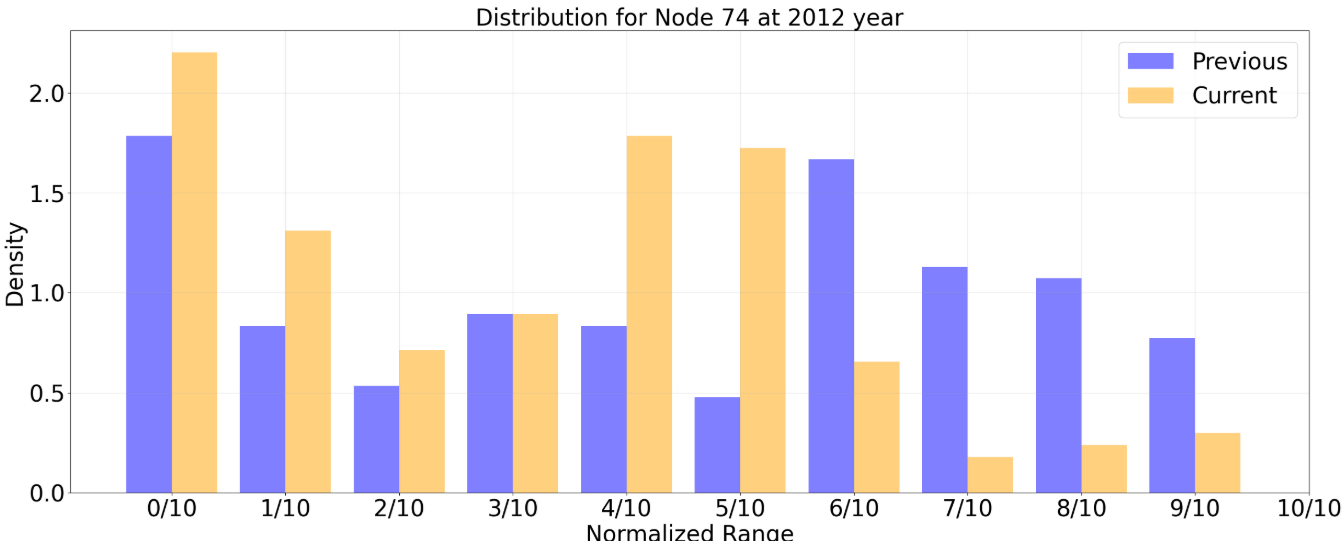}
    \caption{PEMSD3(S), 2012}
  \end{subfigure}\hfill
  \begin{subfigure}{0.315\textwidth}
    \centering
    \includegraphics[width=\linewidth]{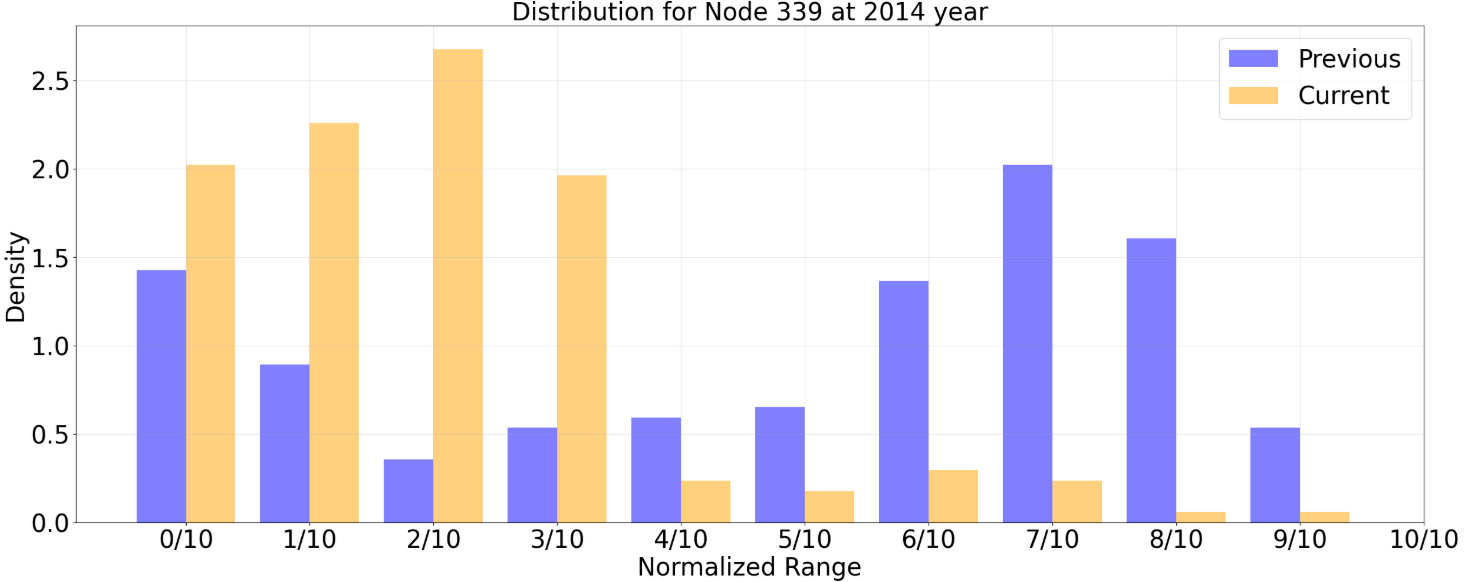}
    \caption{PEMSD3(S), 2014}
  \end{subfigure}\hfill
  \begin{subfigure}{0.315\textwidth}
    \centering
    \includegraphics[width=\linewidth]{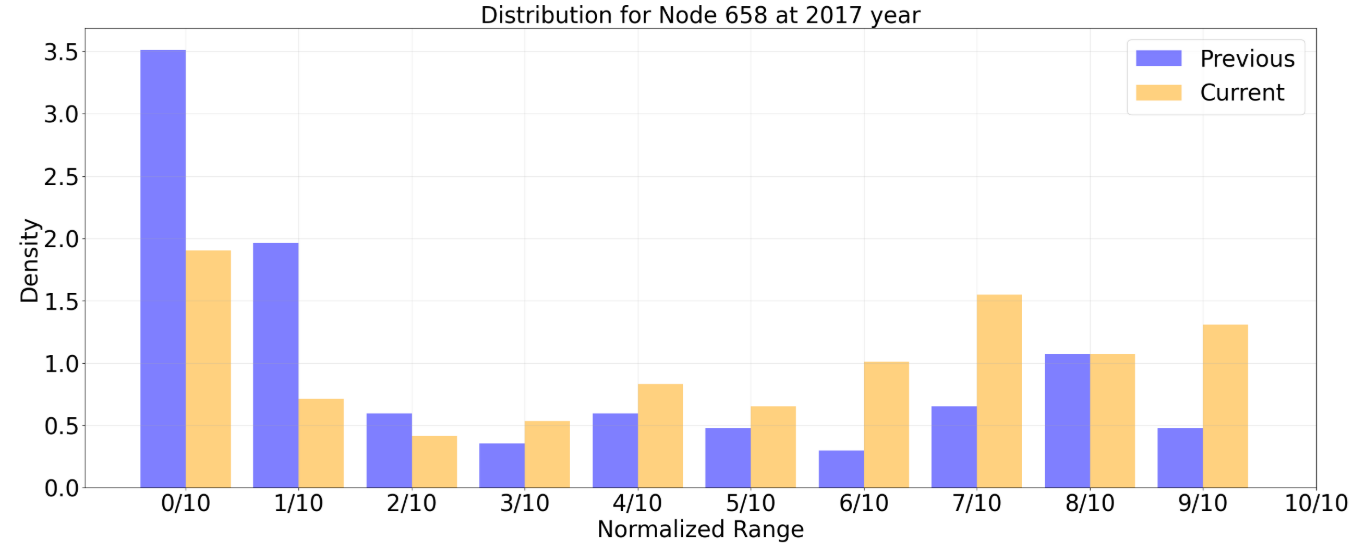}
    \caption{PEMSD3(S), 2017}
  \end{subfigure}
  \vspace{-2mm}

  \begin{subfigure}{0.315\textwidth}
    \centering
    \includegraphics[width=\linewidth]{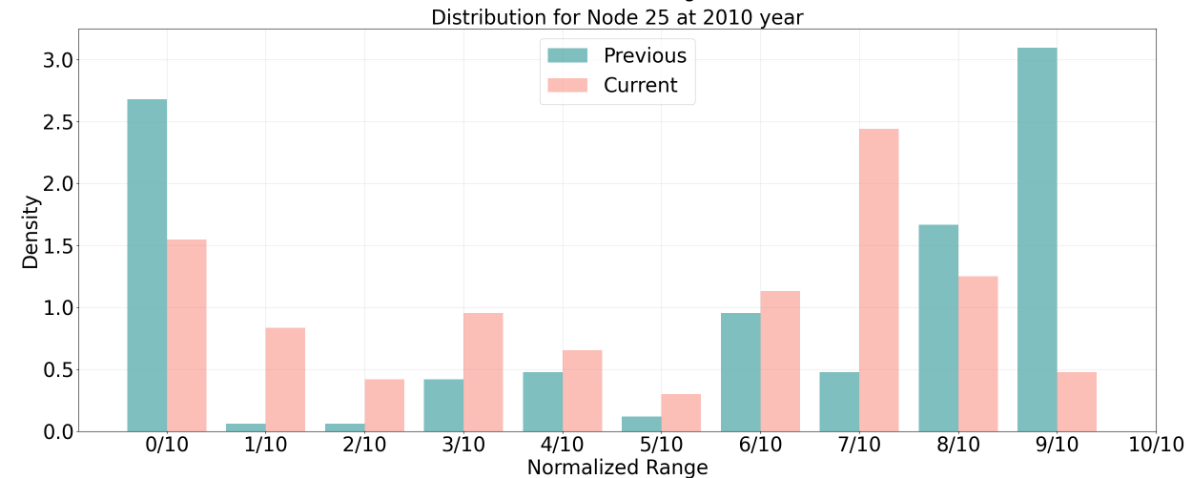}
    \caption{PEMSD4(L), 2010}
  \end{subfigure}\hfill
  \begin{subfigure}{0.315\textwidth}
    \centering
    \includegraphics[width=\linewidth]{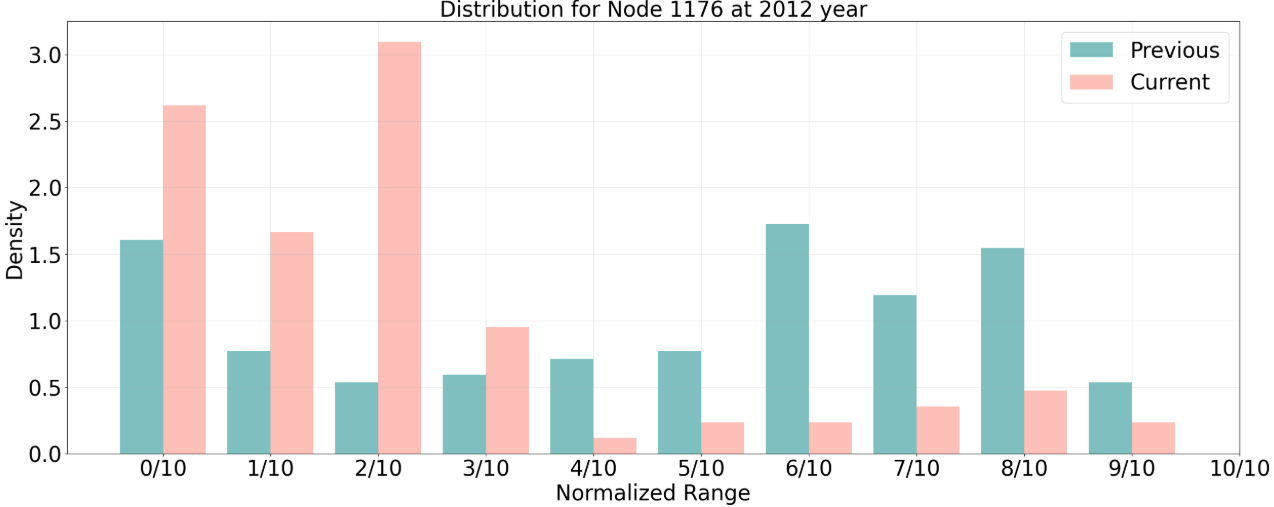}
    \caption{PEMSD4(L), 2012}
  \end{subfigure}\hfill
  \begin{subfigure}{0.315\textwidth}
    \centering
    \includegraphics[width=\linewidth]{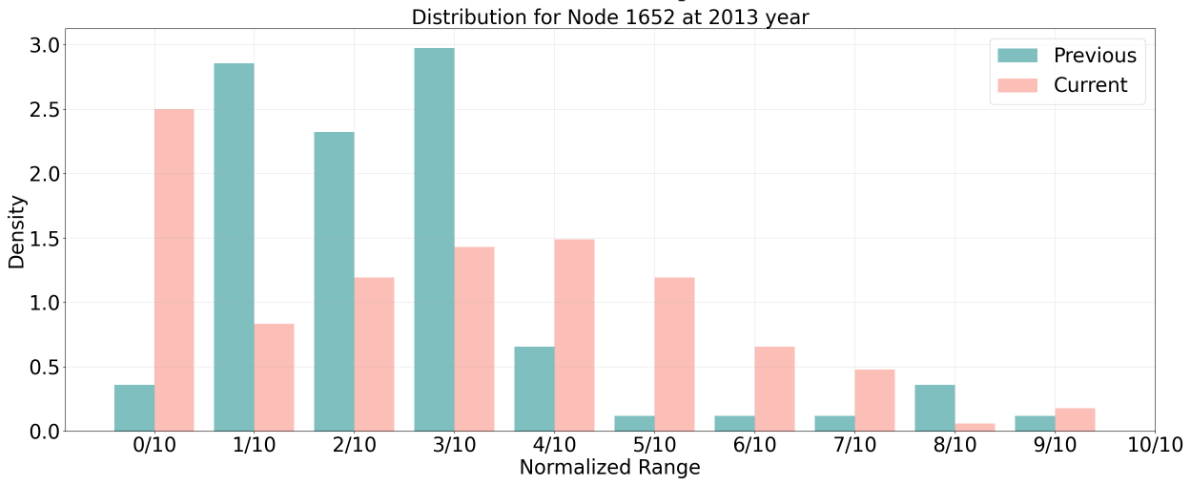}
    \caption{PEMSD4(L), 2013}
  \end{subfigure}
  \vspace{-2mm}

  \begin{subfigure}{0.315\textwidth}
    \centering
    \includegraphics[width=\linewidth]{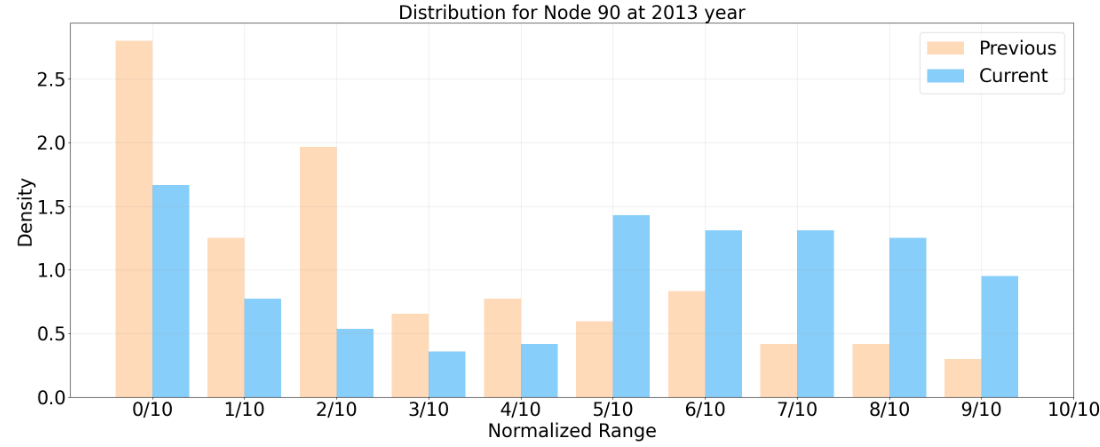}
    \caption{PEMSD8(M), 2013}
  \end{subfigure}\hfill
  \begin{subfigure}{0.315\textwidth}
    \centering
    \includegraphics[width=\linewidth]{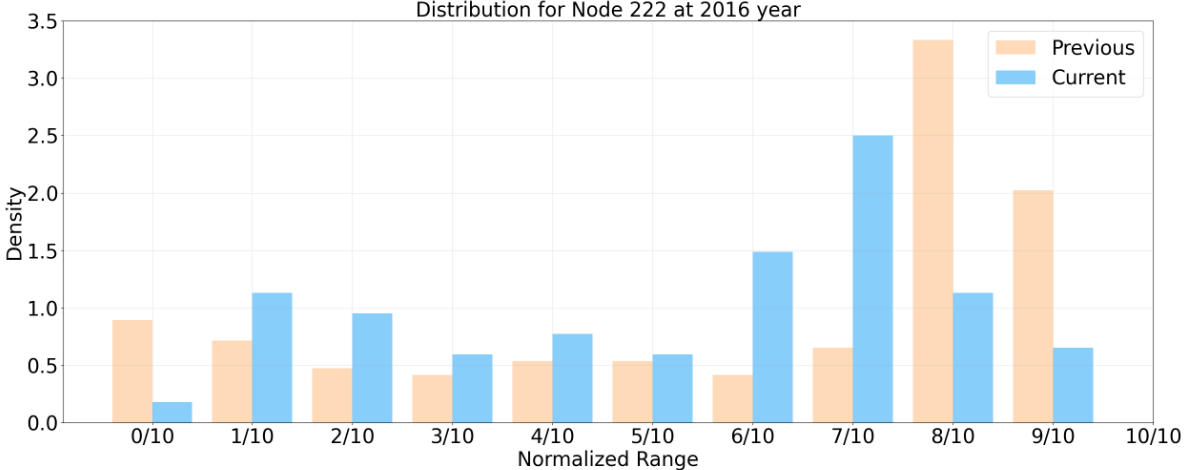}
    \caption{PEMSD8(M), 2016}
  \end{subfigure}\hfill
  \begin{subfigure}{0.315\textwidth}
    \centering
    \includegraphics[width=\linewidth]{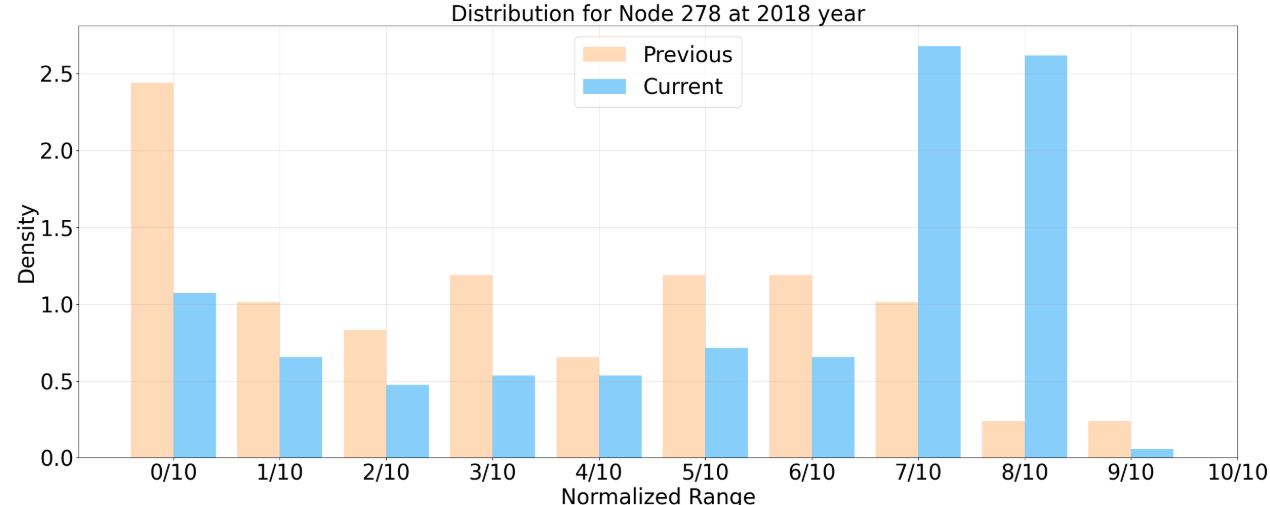}
    \caption{PEMSD8(M), 2018}
  \end{subfigure}
  \caption{Representative normalized empirical feature distributions across three datasets and periods. The examples illustrate heterogeneous distribution shifts: some nodes remain stable, whereas others move between low and high feature-value bins.}
  \label{fig:distribution_visualization}
\end{figure*}

The controlled studies below rely on scalar drift scores. To make the score behavior more interpretable, Fig.~\ref{fig:distribution_visualization} shows representative empirical feature distributions across different periods and datasets. We use this visualization only as qualitative support: the quantitative sampler evidence is reported in Fig.~\ref{fig:sampler_controls} and the attribution analysis discussed above.

Fig.~\ref{fig:distribution_visualization} visualizes the normalized empirical feature distributions defined in Eqs.~\ref{equation2}--\ref{equation3}. The examples support drift-aware allocation of the update budget instead of uniformly treating all shared nodes.

\subsection{Scalability and Controlled Analysis}
Unless otherwise stated, controlled studies use PEMSD3(S), seed 0, the same backbone, split, MAE objective, and drift-selected shared-node budget. Final CoMemNet results use seeds 0--2 and report mean$\pm$std. This section groups the additional revision experiments by the question they answer: whether CoMemNet remains competitive with strong baselines, whether the sampler behavior is explained by simpler heuristics, how the method trades current adaptation against forgetting, and what cost boundary the continual protocol supports.

\subsubsection{Retrained Baselines and Stability}
The STID-retrained and EAC revision runs are merged into Table~\ref{tab:main_results} to keep the main accuracy evidence in one place. STID is a static retraining reference, and EAC is a continual baseline. The separate resource comparison in Table~\ref{tab:recent_static_pemsd3} uses a different batch-size and timing protocol and is therefore reported independently.
\subsubsection{Sampler and Update-Policy Controls}
\begin{figure}[t]
\centering
\includegraphics[width=.98\linewidth]{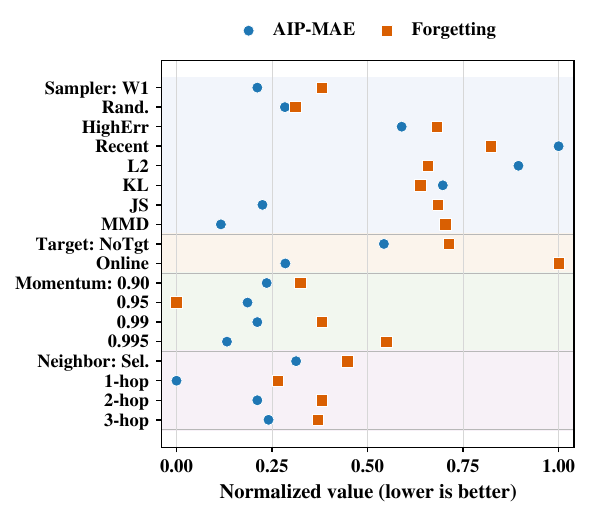}
\caption{Controlled studies on sampler and update policy on PEMSD3(S). AIP-MAE and forgetting are min--max normalized separately for visualization; lower values are better.}
\label{fig:sampler_controls}
\end{figure}

\textbf{The drift sampler performs informative node selection.} Fig.~\ref{fig:sampler_controls} summarizes sampler, Target, momentum, and neighborhood controls under the same protocol. Random tests whether any budgeted subset is sufficient, High-error tests whether recent prediction error alone explains the selected nodes, Recency tests whether the newest windows dominate, and L2/KL/JS/MMD test whether the result depends on a particular distance family. Strict W1 is competitive but not universally best.

\textbf{The sampler mainly reflects traffic-distribution changes rather than graph-structure changes.} Its scores correlate with traffic mean/std./L1 changes ($\rho=0.640,0.573,0.311$), but not with degree/adjacency/neighbor changes ($\rho=-0.032,-0.047,-0.032$). This is consistent with our design choice: the prediction backbone does not use graph structure, while adjacency metadata is used only for optional update-set expansion. Target and neighborhood effects are also modest and mixed; hence these modules are practical update-policy components rather than individually dominant accuracy sources.

\subsubsection{Continual-Learning Metric Matrix}
\begin{table}[t]
\centering
\caption{Continual-learning controls on PEMSD3(S).}
\label{tab:continual_controls}
\resizebox{.45\textwidth}{!}{
\begin{tabular}{lrrrr}
\toprule
Method & Current MAE & AIP & BWT & Forgetting\\
\midrule
Full & 13.706 & 31.652 & -28.015 & 24.013\\
No-replay (no latent state) & 14.419 & 30.324 & -24.199 & 20.742\\
No-TMRB-N & 14.863 & 30.955 & -26.091 & 22.364\\
CoMemNet-retrained & 13.910 & 32.073 & -28.720 & 24.617\\
\bottomrule
\end{tabular}}
\end{table}

Let $R_{i,j}$ be the test MAE on period $j$ after training through period $i$ ($i\ge j$), evaluated on the original test split and node set. We report
\begin{equation}
\begin{aligned}
\mathrm{AIP} &= \frac{2}{T(T+1)}\sum_{i=1}^{T}\sum_{j=1}^{i} R_{i,j},\\
\mathrm{BWT} &= \frac{1}{T-1}\sum_{j=1}^{T-1}(R_{j,j}-R_{T,j}),\\
\mathrm{Forgetting} &= \frac{1}{T-1}\sum_{j=1}^{T-1}\left(R_{T,j}-\min_{i\in\{j,\dots,T\}}R_{i,j}\right).
\end{aligned}
\end{equation}
Lower AIP and Forgetting are better, larger BWT is better, and periods are weighted equally. Table~\ref{tab:continual_controls} separates adaptation to the latest period from historical performance variation. Full CoMemNet achieves the best Current MAE by updating newly added and drift-sensitive shared nodes while reusing compact temporal states. Compared with CoMemNet-retrained and the two ablated variants, it improves current-period adaptation and keeps AIP, BWT, and Forgetting within a comparable range. This supports the intended role of CoMemNet: efficient adaptation to an evolving road network under a bounded update protocol, rather than optimizing a single forgetting metric in isolation.

\subsubsection{Budget, Robustness, and Resource Accounting}
These checks define the operating boundary of CoMemNet rather than introducing additional model components. Table~\ref{tab:rho_sensitivity} shows the effect of the selected-node budget: larger budgets generally improve current-period MAE but also increase update cost and historical interference, so $\rho$ is treated as a deployment-dependent stability--plasticity parameter. Robustness is evaluated with a fixed 2017 checkpoint; Full CoMemNet changes from 15.324 clean MAE to 15.524/15.602 under 10\%/20\% missing sensors and 16.333 under 10\% noise, remaining better than No-replay under the same corruptions. The resource evidence is reported in Table~\ref{tab:main_results}, Table~\ref{tab:recent_static_pemsd3}, and Fig.~\ref{fig:round2_scaling_advantage}.

\section{Conclusion}
We presented CoMemNet, a Continual Memory Network with drift-aware sampling for evolving traffic sensors. The prediction backbone is adjacency-free, while adjacency metadata may assist update-set expansion. An EMA Target branch supports Wasserstein-1 representation-discrepancy ranking, and TMRB-N stores compact temporal states for latent-state reuse. Three-seed results, official baselines, sampler controls, robustness tests, and resource accounting support efficient continual adaptation and a practical stability--plasticity trade-off, rather than topology causality or universal forgetting mitigation. Event-driven topology changes and stronger nonstationary shocks remain future work.

\section*{Acknowledgment}
This work was supported by the China Association for Science and Technology (CAST) Young Talents Cultivation Program.

\bibliographystyle{ieeetr}
\bibliography{sample-base}

\begin{IEEEbiography}[{\includegraphics[width=1in,height=1.25in,clip,keepaspectratio]{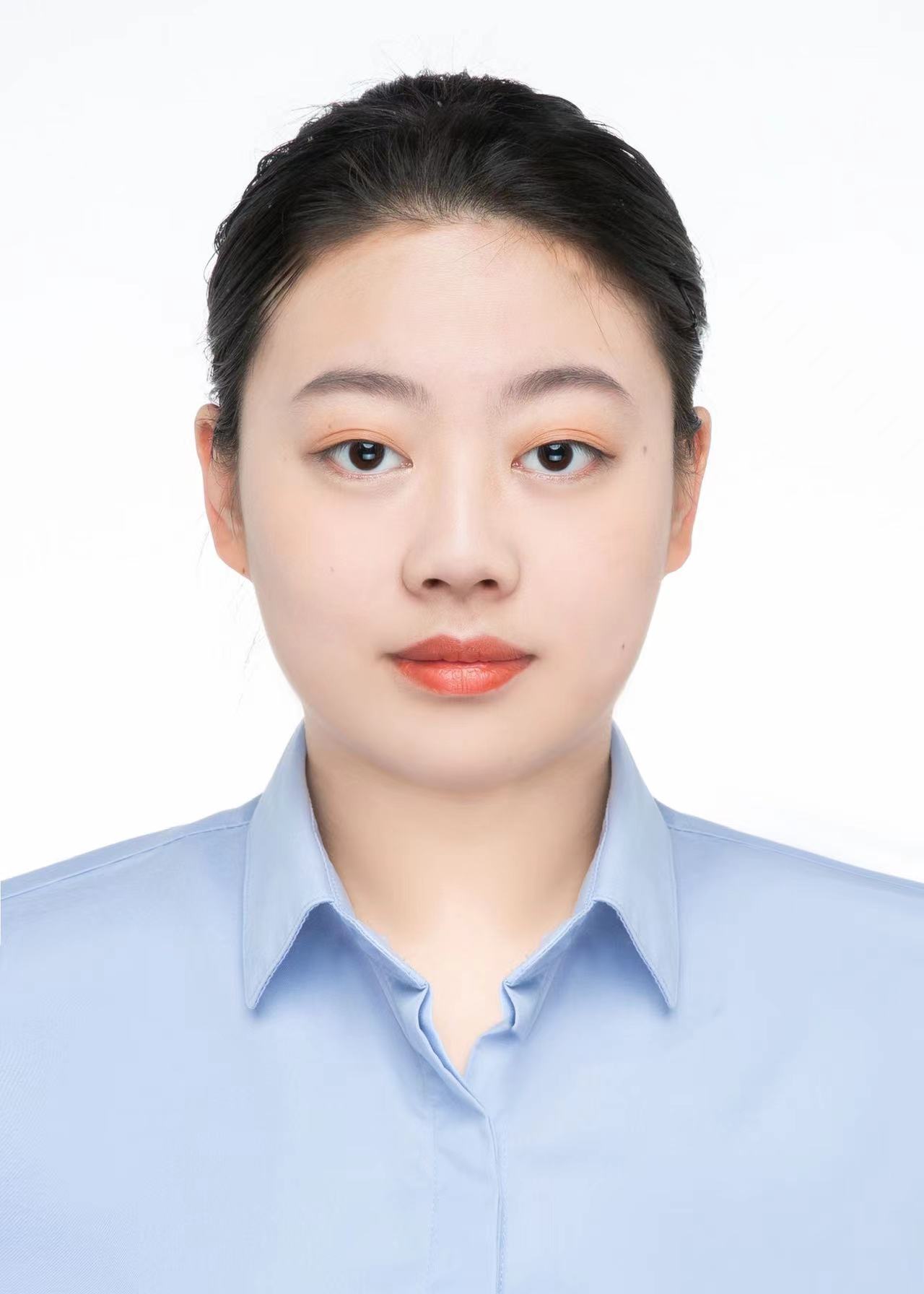}}]{Mei Wu}{\space}is currently pursuing the Ph.D. degree at Shanghai Jiao Tong University, China. Her research interests include spatiotemporal data mining and traffic prediction.
\vspace{-6mm}
\end{IEEEbiography}

\begin{IEEEbiography}[{\includegraphics[width=1in,height=1.25in,clip,keepaspectratio]{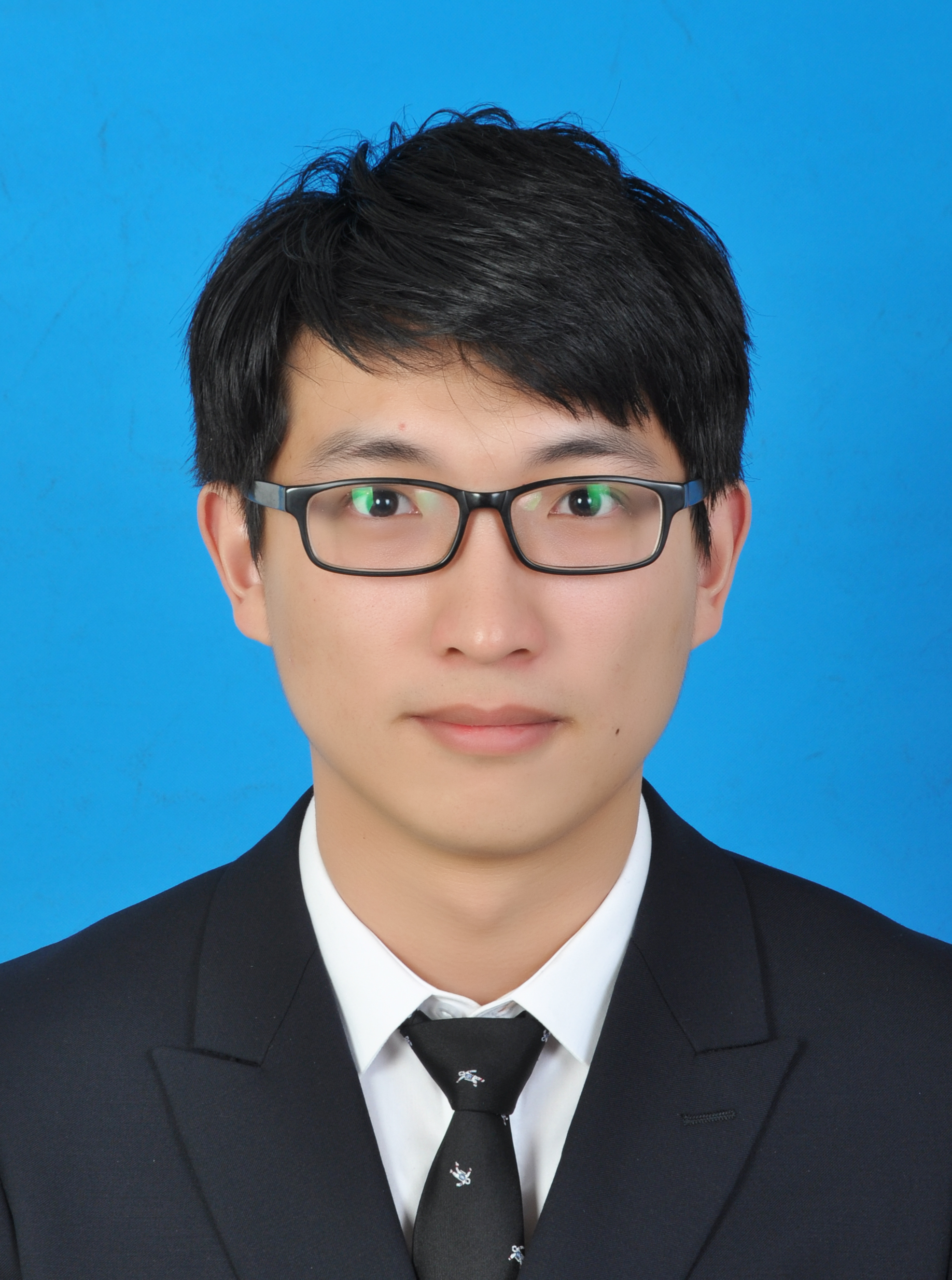}}]{Wenchao Weng}{\space}is currently pursuing the Ph.D. degree at Zhejiang University of Technology, China. His research interests include spatiotemporal graph learning and traffic forecasting.
\vspace{-6mm}
\end{IEEEbiography}

\begin{IEEEbiography}[{\includegraphics[width=1in,height=1.25in,clip,keepaspectratio]{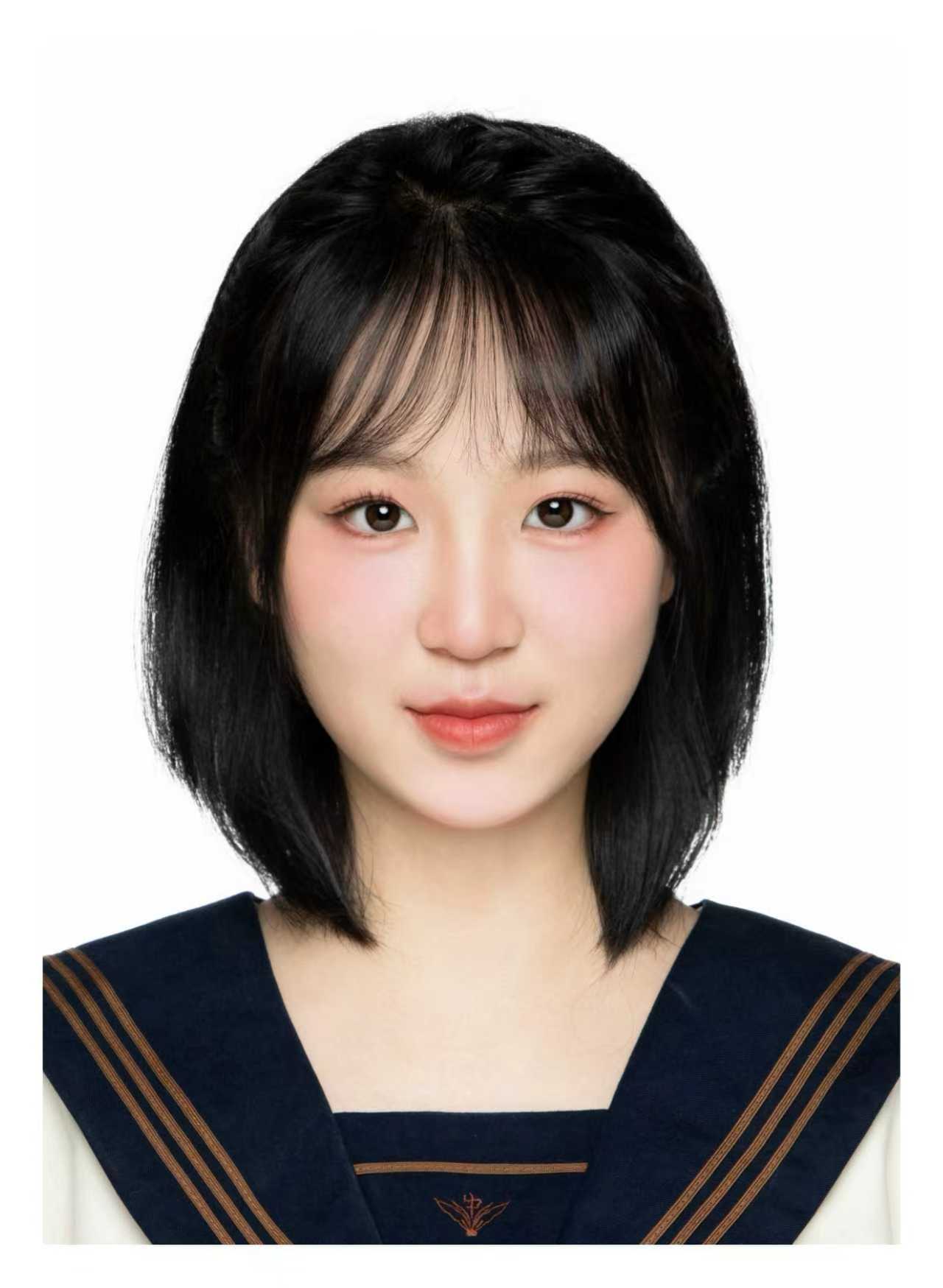}}]{Wenxin Su}{\space}is currently pursuing the B.S. degree at Zhejiang University of Technology, China. Her research interests include deep learning and spatiotemporal prediction.
\vspace{-6mm}
\end{IEEEbiography}

\begin{IEEEbiography}[{\includegraphics[width=1in,height=1.25in,clip,keepaspectratio]{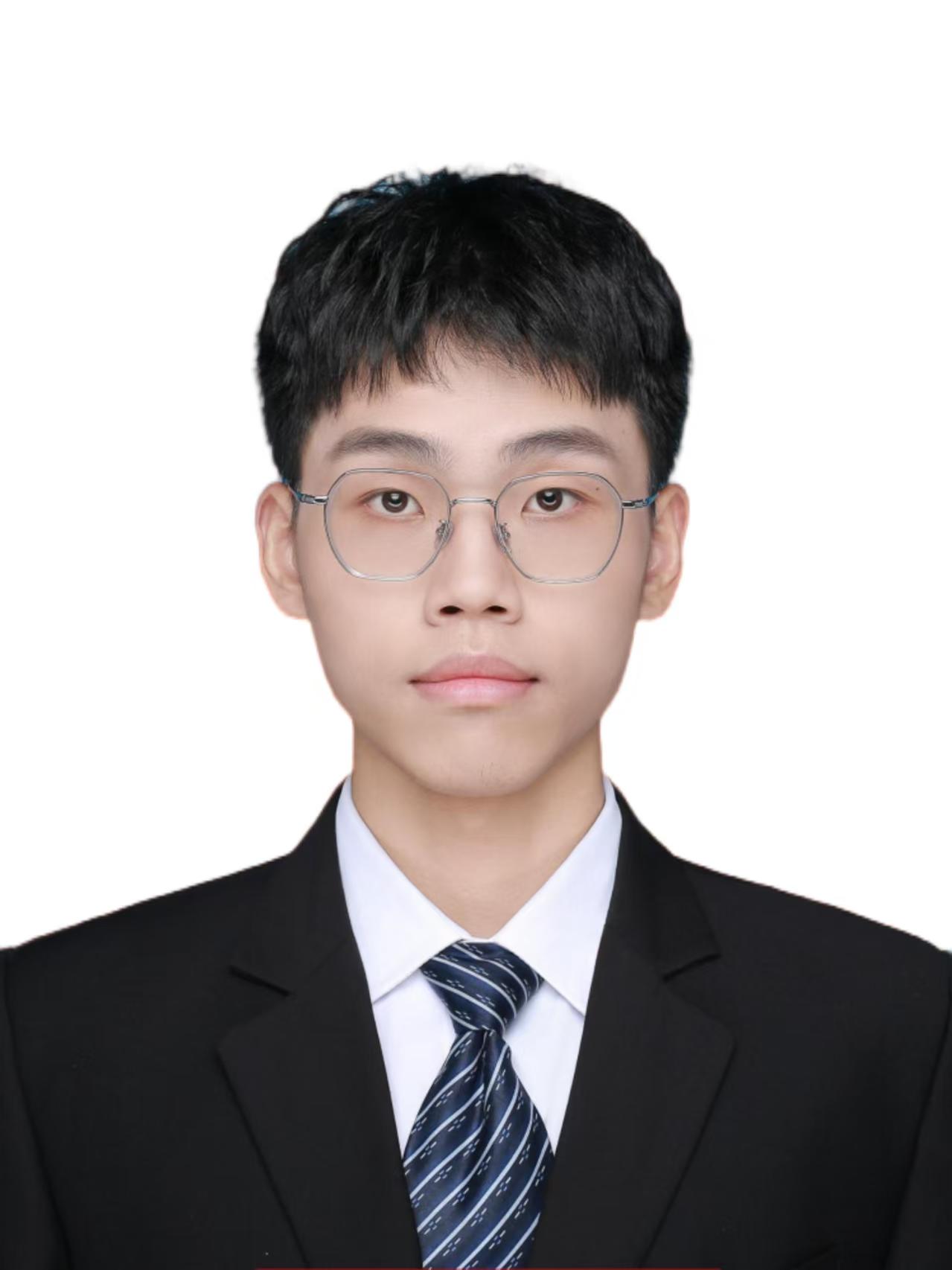}}]{Wenjie Tang}{\space}is currently pursuing the B.S. degree at Nanjing University of Science and Technology, China. His research interests include robotics and human--robot interaction.
\vspace{-6mm}
\end{IEEEbiography}

\begin{IEEEbiography}[{\includegraphics[width=1in,height=1.25in,clip,keepaspectratio]{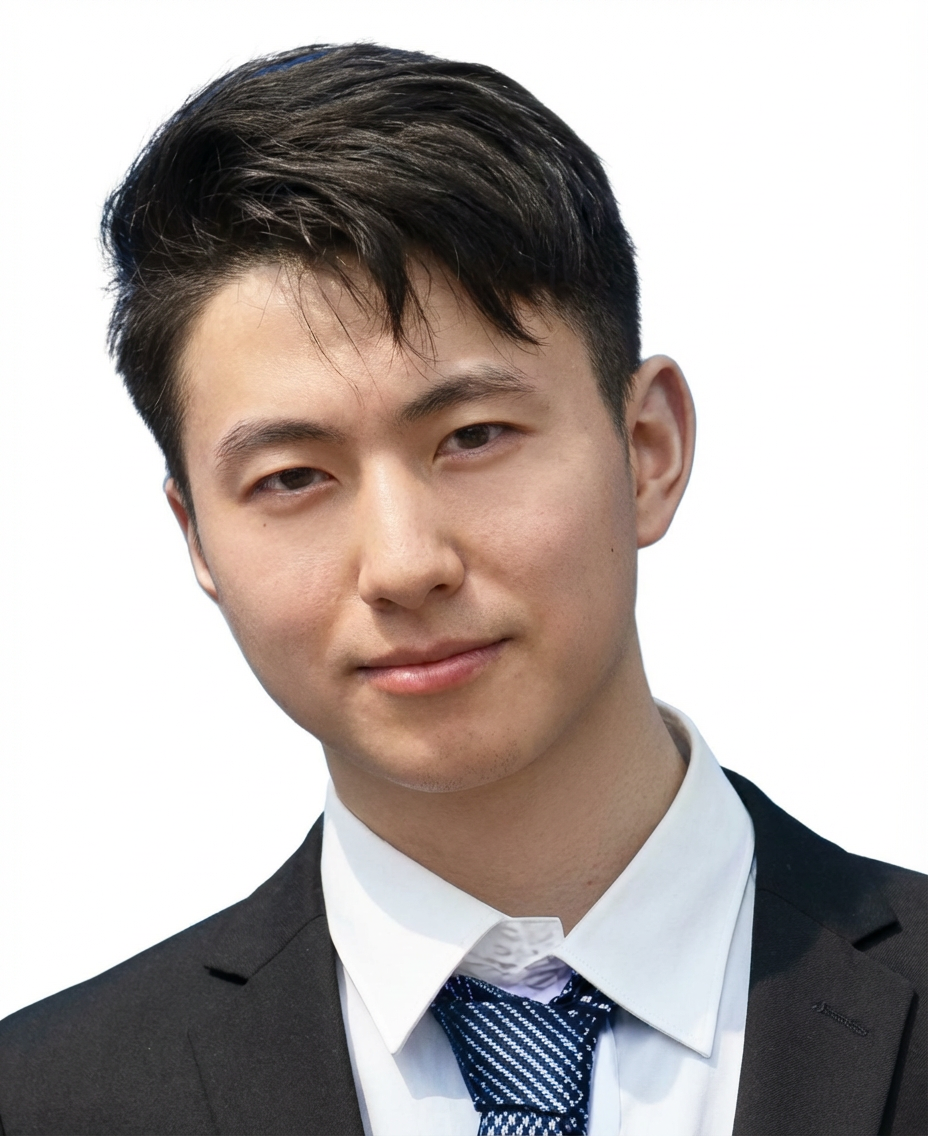}}]{Wei Zhou}{\space}is an Associate Professor at Nanjing University of Science and Technology, China. His research interests include computer vision, robotics, and intelligent transportation.
\end{IEEEbiography}

% \begin{IEEEbiography}[{\includegraphics[width=1in,height=1.25in,clip,keepaspectratio]{xiafeng.jpg}}]{Feng Xia}{\space}(Senior Member, IEEE) received the BSc and PhD degrees from Zhejiang University, Hangzhou, China. He is a Professor in School of Computing Technologies, RMIT University, Australia. Dr. Xia has published over 300 scientific papers in journals and conferences (such as IEEE TAI, TKDE, TNNLS, TC, TMC, TBD, TCSS, TNSE, TETCI, TETC, THMS, TVT, TITS, TASE, ACM TKDD, TIST, TWEB, TOMM, WWW, AAAI, ICLR, SIGIR, WSDM, CIKM, JCDL, EMNLP, and INFOCOM). His research interests include artificial intelligence, graph learning, brain science, digital health, and robotics. He is a Senior Member of IEEE and ACM, and an ACM Distinguished Speaker.

% %\vadjust{\vfill\pagebreak} 用来换页
% \end{IEEEbiography}

\end{document}